\documentclass{article}

\PassOptionsToPackage{numbers, compress}{natbib}
\usepackage[final,main]{neurips_2026}

\usepackage[utf8]{inputenc}
\usepackage[T1]{fontenc}
\usepackage{hyperref}
\usepackage{url}
\usepackage{booktabs}
\usepackage{amsfonts}
\usepackage{amsmath}
\usepackage{amssymb}
\usepackage{amsthm}
\usepackage{nicefrac}
\usepackage{microtype}
\usepackage{xcolor}
\usepackage{graphicx}
\usepackage{multirow}
\usepackage{makecell}
\usepackage{array}
\usepackage{enumitem}
\usepackage{wrapfig}
\usepackage{algorithm}
\usepackage{algpseudocode}
\graphicspath{{fig/}{../figs/pdf/}{figs/pdf/}}

\newcommand{\teacher}{f_{\theta}}
\newcommand{\student}{f_{\tilde\theta}}

\newcommand{\Lkl}{\mathcal{L}_{\text{KL}}}
\newcommand{\Lce}{\mathcal{L}_{\text{CE}}}
\newcommand{\Lheed}{\mathcal{L}_{\text{HEED}}}

\newcommand{\Lalign}{\mathcal{L}_{\text{align}}}
\newcommand{\Lrsa}{\mathcal{L}_{\text{RSA}}}
\newcommand{\Lkd}{\mathcal{L}_{\text{KD}}}
\newcommand{\rorig}{r^{\theta}}
\newcommand{\rconv}{r^{\tilde\theta}}

\newtheorem{assumption}{Assumption}

\newtheorem{remark}{Remark}

\title{HEED: Density-Weighted Residual Alignment for \\ Hybrid Vision-Language Model Distillation}

\author{%
  Yihao Liang \qquad Niraj K. Jha \\
  Department of Electrical and Computer Engineering \\
  Princeton University, Princeton, NJ 08544, USA \\
  \texttt{\{yhliang, jha\}@princeton.edu}
}

\begin{document}

\maketitle

\begin{abstract}
Distilling vision-language models into faster hybrid architectures, such as 3:1 Mamba-2/attention mixes, is now standard practice for making inference efficient. Aggregate benchmarks suggest that this works but they hide selective failures. When we distill Qwen3-VL-8B-Instruct into a 3:1 Mamba-2/attention hybrid, student model stays within 2 points of the teacher across visual reasoning benchmarks like MMStar, MMBench, and MMMU-Pro, while dropping 13 points on optical-character-recognition and document tasks. The student can still understand the scene but loses the fine-grained text needed to answer. 
We localize much of the failure to a specific kind of position. In a high-resolution image, most patches are sky, wall, or smooth texture, while a small fraction carries text, edges, object boundaries, or other local details. In a token-level diagnostic, the top 10\% highest-density patches have 3.6$\times$ larger residual drift than the bottom 10\% lowest-density patches and 3.5$\times$ larger teacher-masking answer contribution. Uniform weighting devotes many loss terms to low-information background patches, whereas sparse answer-bearing patches receive no special protection.
The required intervention is minimal: we replace uniform residual alignment with density-weighted residual alignment, using patch self-dissimilarity as a training-free proxy for position importance. We call this HEED. Compared with normal end-to-end distillation, HEED increases performance by 8.7 points on OCRBench~v2 and 5.13 points on a 10-benchmark average. The gain is realized on different teacher models. After standard post-training, the student reaches teacher-level performance on the 10-benchmark average. HEED adds no parameters and no inference-time cost. Hence, the student maintains the hybrid architecture's 4.12$\times$ throughput and 68\% memory saving at 128k context. Code is available at \url{https://github.com/jha-lab/HEED}.
\end{abstract}

\section{Introduction}
\label{sec:intro}

Hybrid vision-language models (VLMs) are now a standard answer to slow inference: maintain a small amount of attention, replace most layers with a linear-time mixer, and serve longer multimodal contexts at a fraction of the cost~\cite{lieber2024jamba,jetnemotron2025,nvidia2025nemotronh,qwen2026qwen35,qwen2026qwen36}. Training such models from scratch remains expensive. Hence, knowledge distillation (KD) from a pretrained VLM is a practical alternative. On aggregated benchmarks, this recipe looks fine. However, it remains unclear as to which multimodal capabilities are most fragile.

The aggregate hides some information. When we distill Qwen3-VL-8B-Instruct~\cite{qwen3vl2025} into a 3:1 Mamba-2 hybrid using standard KD pipelines, the student stays within about 2 points of the teacher on MMStar~\cite{mmstar}, MMBench~\cite{mmbench}, and MMMU-Pro~\cite{mmmu_pro} benchmarks but exhibits a loss of 13 points on OCRBench~v2~\cite{ocrbenchv2} and InfoVQA~\cite{infovqa} benchmarks. For example, the student can still describe that an image contains a receipt but misreads the digits on it. Aggregate scores miss this failure. Prior hybrid VLMs exhibit the same asymmetry: the performance of mmMamba's TextVQA drops by 18.3 points where that of its POPE by only 2.4 points~\cite{mmmamba2025}. Yet, prior work has not explained why the failure mode is so selective.

\begin{figure}[t]
\centering
\includegraphics[width=\linewidth]{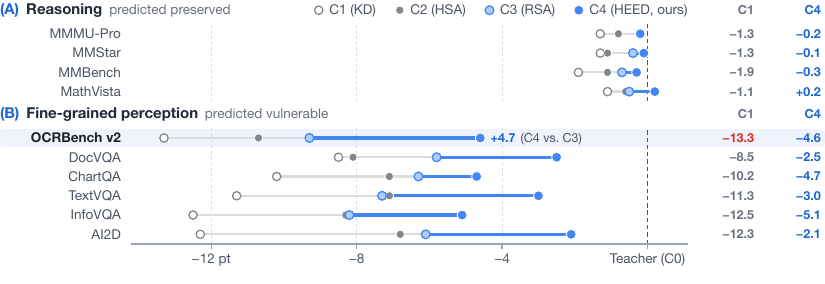}
\vspace*{-8mm}
\caption{\textbf{Aggregate scores hide a fine-grained perception collapse that HEED partially recovers from.} \textit{(A)} Reasoning benchmarks (predicted preserved): C1-C4 stay within $\sim$2~pt of the teacher. \textit{(B)} Fine-grained perception (predicted vulnerable): standard KD (C1) loses 13.3~pt on OCRBench~v2; density weighting (C4 vs.\ C3) recovers +4.7~pt under matched setup. C1--C4 share architecture and data; only the Stage~1/2 alignment loss differs. Dots show each condition's score gap to the teacher.}

\vspace*{-6mm}
\label{fig:teaser}
\end{figure}

Fig.~\ref{fig:teaser} shows the benchmark pattern with five matched conditions: C0: Teacher (baseline), C1: standard end-to-end KD, C2: hidden-state alignment (HSA), C3: uniform residual-stream alignment (RSA), and C4: HEED. C3 is our internal control: It uses the same staged recipe as C2 but aligns the residual stream instead of per-layer block outputs. Tab.~\ref{tab:main} lists the per-benchmark scores.

Sect.~\ref{sec:diagnostic} describes how to localize the failure: A small fraction of visual patches contains the key information  (digits, labels, chart marks, edges) needed to answer the question and the hybrid student drifts disproportionately on those locally distinctive, \emph{high-density} positions, defined by the Vision Transformer (ViT)~\cite{dosovitskiy2020image} feature differing from nearby patches. The top 10\% patches are 3.6$\times$ farther from the teacher than the bottom 10\% and contribute 3.5$\times$ more to the teacher's answer when masked. A random control rules out the simpler explanation that any extra-weighted subset would do: The extra weight has to be assigned to the high-density patches.

Our method, HEED, implements the corresponding intervention without changing the model. It computes density weights once from  ViT features and uses them only during residual alignment in KD. While using the same architecture, data, and budget, HEED increases performance by  8.7 points on OCRBench~v2 and 5.13 on the 10-benchmark average relative to normal end-to-end KD (C1). In the C3$\to$C4 comparison, where only the per-position residual-alignment weight changes, the gain is 4.7 and 2.24 points, respectively. Unlike inference-time visual-token reduction, HEED keeps all tokens during inference and only reweights the alignment loss across positions during the teacher-to-hybrid conversion. Hence, these two directions are  complementary. After standard supervised finetuning (SFT) and direct preference optimization (DPO)~\cite{dpo2023}, the student reaches teacher-level performance on the 10-benchmark average. The 4.1$\times$ throughput and 68\% memory saving at 128k context come from the hybrid architecture itself.

\vspace*{-4mm}
\paragraph{Contributions.}
\begin{enumerate}[leftmargin=1.4em,topsep=2pt,itemsep=1pt]
\item \emph{Diagnosis.} We show that hybrid VLM distillation can preserve reasoning while losing the visual details needed for OCR and document understanding 
(Sect.~\ref{sec:diagnostic}).
\vspace*{-1mm}
\item \emph{Mechanism.} We show that high-density visual positions drift most from the teacher and matter most for the teacher's answer when masked. A random control shows that the benefit does not come from giving extra weight to arbitrary patches. The extra weight needs to be assigned to high-density patches
(Sect.~\ref{sec:diagnostic}).
\vspace*{-1mm}
\item \emph{Method and result.} HEED improves performance by 8.7 points on OCRBench~v2 and 5.13 on average over end-to-end KD, without adding parameters and at no inference-time cost. This improvement endures over different seeds and baseline VLMs. HEED also trains a Gated DeltaNet (GDN) hybrid
(Sect. \ref{sec:method}, 
\ref{sec:experiments}).
\end{enumerate}

These results highlight a broader principle: When heterogeneous inputs are compressed into a fixed-capacity hybrid sequence model, distillation should not treat all tokens uniformly but instead should reflect their varying importance.
\section{Related work}
\label{sec:related}

Linear architectures, such as Mamba, Mamba-2, and GDN, reduce long-sequence modeling cost from quadratic to linear~\cite{gu2023mamba,dao2024mamba2,yang2025gateddeltanet} but their performance is weaker than attention at exact token recall tasks. Hybrid stacks are a practical compromise: Most layers use a linear-time mixer for efficiency, whereas periodic full-attention layers preserve precise token interactions~\cite{Waleffe2024AnES,qwen2026qwen35}.
Training a large hybrid model from scratch, however, is expensive. Therefore, most work starts from a pretrained Transformer and distills it into a hybrid architecture. Existing recipes include MOHAWK~\cite{bick2024mohawk}, Mamba-in-Llama~\cite{wang2024mambainllama}, LoLCATs~\cite{zhang2024lolcats}, RADLADS~\cite{goldstein2025radlads}, and Jet-Nemotron~\cite{jetnemotron2025}. These methods differ in initialization and training schedule but their alignment losses share a common assumption: Every token or visual patch contributes to the distillation based on the same weight. Our results show that this assumption is fragile for multimodal sequences, where a blank background patch and a receipt digit should not be equally protected.

The problem is especially visible in VLMs. MaTVLM~\cite{matvlm2025} and mmMamba~\cite{mmmamba2025} are the recent cross-architecture distillation methods for VLMs. However, they do not directly address fine-grained grounding and OCR-style failures. The evidence they report suggests a selective failure: the performance of mmMamba drops by 10.9 points on TextVQA with $\sim$75\% Mamba-2 layers and 18.3 points at 100\% but only by 0.9 and 2.4 points, respectively, on POPE. We investigate this asymmetry as the starting point, measure it at the token level, and ask which visual positions are most damaged by KD.

HEED addresses the above problem through loss weighting. Previous weighting schemes usually reweight \emph{examples} (focal loss~\cite{lin2017focal}, hard-example mining), \emph{logits} (KD temperature~\cite{hinton2015distillation}), or \emph{task losses} (multi-task balancing). Attention transfer~\cite{zagoruyko2017at} and token-level KD in language models provide the closest precedents for per-position alignment weighting. Yet, they leave unaddressed a key question for hybrid VLMs: Which visual positions should be preserved under limited state capacity?
Saliency-aware methods derive importance from teacher attention, class-activation maps such as Grad-CAM~\cite{selvaraju2017gradcam}, or gradients, and use them to weight logits, losses, or feature maps. HEED weights residual-stream positions at the replaced layers with frozen input features and needs no labels or teacher backward pass. Local self-dissimilarity itself is a classical descriptor~\cite{shechtman2007selfsim}. Our contribution is conversion diagnosis and the finding that this training-free signal recovers most of the gradient-weighted benefit (Appendix~\ref{app:importance-aware}).
Classical importance weighting~\cite{shimodaira2000importance,hampel1974influence} provides a simple intuition: Mistakes at more important positions should count more during training. HEED applies this idea to the residual stream by using patch self-dissimilarity as an efficient estimate of per-position importance and shows that it agrees with a more expensive gradient-based measure.

HEED is complementary to inference-time VLM efficiency methods, such as VisionZip, FastV, and VoCo~\cite{visionzip2024,fastv2024,voco2024}, which exploit patch heterogeneity by deciding which tokens to keep at inference time. In contrast, HEED keeps all tokens at inference time and instead reweights the alignment loss across various positions during the teacher-to-hybrid conversion. This targets the conversion bottleneck rather than the inference-cost bottleneck. Therefore, it is complementary to inference-time token-reduction methods. Unlike gradient-based importance methods, HEED estimates position importance using only frozen ViT features, making it teacher-agnostic.

\section{Diagnosing selective failure in hybrid VLM distillation}
\label{sec:diagnostic}

\begin{figure}[t]
\centering
\includegraphics[width=\linewidth]{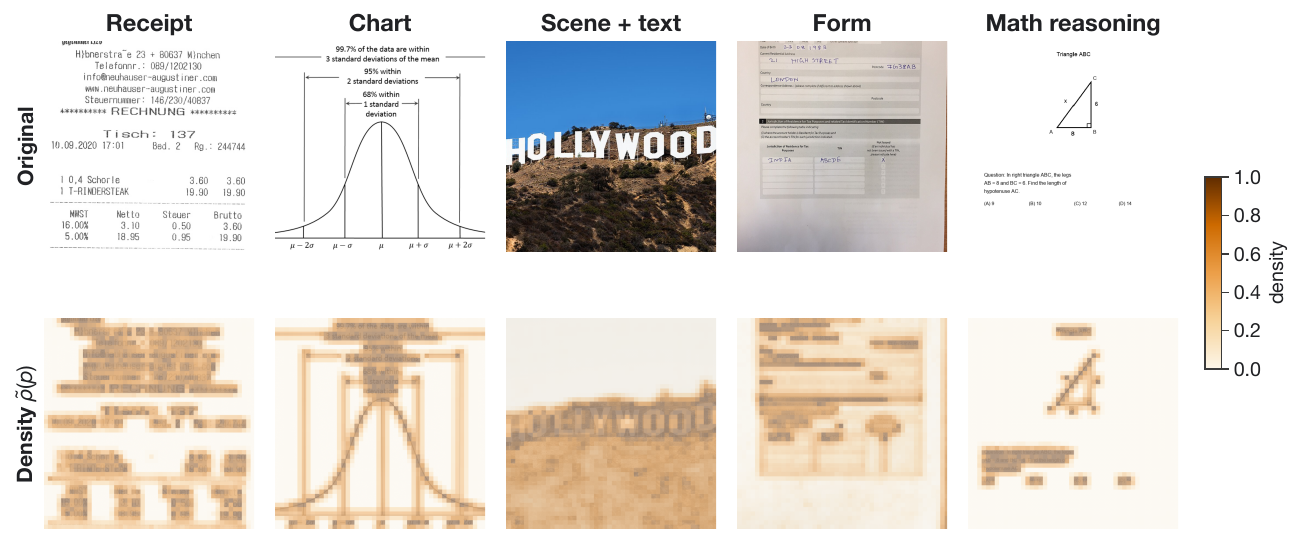}
\vspace*{-2mm}
\caption{\textbf{Example density maps $\tilde{\rho}(p)$ across five visual domains.} 
Dark patches are locally distinctive (high $\tilde{\rho}$); bright patches resemble their neighbors (low $\tilde{\rho}$). The signal highlights text, chart marks, form fields, signs, and small labels rather than smooth background regions, without using any task supervision. Visualization uses CLIP ViT-L/14 features for clarity; the training pipeline applies the same statistic (Eq.~\ref{eq:density}) to Qwen3-VL's frozen ViT features.}
\vspace*{-3mm}
\label{fig:density-examples}
\end{figure}

After regular KD, the student (C1) can often recognize a scene, e.g., that an image contains a receipt, chart, or form, but loses the small details needed to provide an answer: digits, labels, fine text, and marks. The diagnostic discussed in this section tests a specific prior. Fine-grained perception may depend on a small fraction of locally distinctive visual positions, while reasoning may depend on more broadly distributed token-level evidence. If this prior holds, a fixed-capacity hybrid should drift more on the concentrated visual positions; therefore, the resulting benchmark damage should be selective. We test four candidate explanations for where residual-stream drift concentrates: \emph{local visual density} (our prediction), \emph{token type} (visual or text), \emph{layer depth}, and \emph{teacher attention concentration}. We use local visual density as an operational proxy for visual information density: A patch has high density when its ViT feature differs from the features in its 3${\times}$3 neighborhood. This is a simple input-side measurement, not a semantic saliency label. Fig.~\ref{fig:density-examples} demonstrates the signal: text characters, chart marks, form fields, signs, and small labels are dark while smooth backgrounds are bright.

\subsection{Diagnostic setup}
\label{sec:diag-setup}

We run the diagnostic on C1, the normal end-to-end KD baseline: Qwen3-VL-8B-Instruct distilled into the 3:1 Mamba-2 hybrid with end-to-end Kullback-Leibler (KL) divergence and cross-entropy (CE) loss. Let $\rorig_{\ell,p}$ and $\rconv_{\ell,p}$ be the teacher and student residual streams at layer $\ell$, position $p$. On 1,000 samples sampled from the held-out validation slice of the LLaVA-OneVision-Data~\cite{li2024llava} subset, we record two token-level measurements:
\[
\delta_{\ell,p}=\|\rconv_{\ell,p}-\rorig_{\ell,p}\|_2,
\]
the residual drift, i.e., how far the hybrid student has moved from the teacher, and $a_{\ell,p}$, the answer importance measured by masking that token in the teacher, i.e., how much the teacher's answer score drops when the token is hidden from attention.

\subsection{Density predicts residual drift}
\label{sec:diag-density}

We first ask which tested factor best predicts residual drift. We fit a token-level linear regression with all four predictors standardized, and read off each factor's unique contribution as its \emph{semi-partial $R^2$} --- the extra fraction of drift variance that this predictor explains after the other three are already in the model (a larger $R^2$ implies greater usefulness). The per-factor unique contributions are density 0.30, token type 0.10, layer depth 0.08, and teacher attention 0.05, with joint $R^2${=}0.53. Density's 95\% confidence interval from a 1{,}000-image bootstrap is [0.27, 0.33] and is the largest in every resample ($p{<}10^{-3}$). It, therefore, explains roughly three times the unique drift variance of the next factor. We read this as a diagnostic decomposition, not a causal proof: The strict causal test, whether density-weighted alignment actually reduces drift, comes from random-position control 
(Sect.~\ref{sec:diag-design}) and C3$\to$C4 intervention 
(Sect.~\ref{sec:exp-main}). The full bootstrap procedure is presented in Appendix~\ref{app:density-decomposition}.

This is not just a visual-vs.-text effect. If the problem were simply that visual tokens fail, token type would remain a strong predictor after density is included. Instead, its unique contribution drops to 0.04-0.06 once density enters the model. The better-fitting reading is local distinctiveness: Text characters, chart marks, form fields, and small visual details drift more when they stand out from their neighbors, whereas smooth visual regions do not drift in the same fashion.

\begin{figure}[t]
\centering
\includegraphics[width=\linewidth]{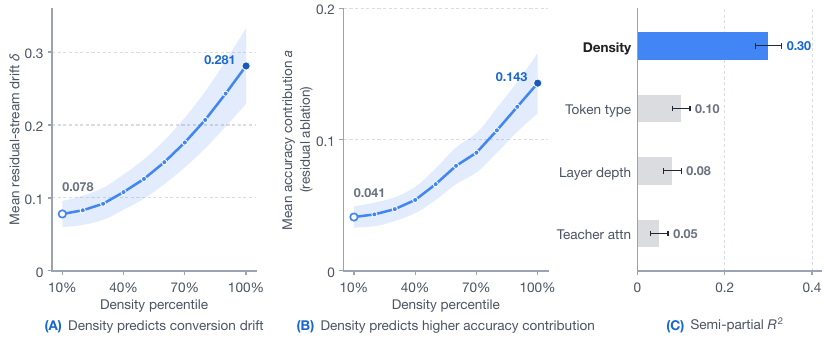}
\caption{\textbf{High-density tokens drift more and matter more for the teacher's answer.} \textit{(A)} Residual-stream drift $\delta$ between C1 student and teacher: top-10\% density tokens drift 3.6$\times$ over bottom-10\%. \textit{(B)} Per-token residual-ablation effect on the teacher's answer score: 3.5$\times$ ratio between top- and bottom-10\%. \textit{(C)} semi-partial $R^2$: Density explains $\sim$3$\times$ more unique variance than token type, layer depth, or teacher attention. Bands show within-group variation across 1{,}000 held-out samples.}
\label{fig:diagnostic}
\end{figure}

The same conclusion be drawn in a simpler top-vs.-bottom comparison. Sort tokens by density into ten equal-size groups: The bottom 10\% has mean residual drift 0.078 vs.\ 0.281 for the top 10\% (3.6$\times$ larger. Fig.~\ref{fig:diagnostic}A). Masking those same groups in the teacher changes the answer score by 0.041 vs.\ 0.143 (3.5$\times$ larger. Fig.~\ref{fig:diagnostic}B). High-density patches are, therefore, both where the hybrid student drifts most and where masking matters most for the teacher's answer. On patches with no detected text, the two ratios remain at 2.91$\times$ and 2.15$\times$ (Appendix~\ref{app:text-control}).

\subsection{From diagnosis to loss design}
\label{sec:diag-design}

One more check provides a simpler explanation: The gain is not just derived from giving extra weight to more positions. When we upweight the same number of positions, choosing them by density beats choosing them at random (Appendix~\ref{app:density-control}). This gives the method its design rule: because residual drift and teacher-masking answer sensitivity are concentrated on high-density positions; thus, those positions should receive more residual-alignment weight. HEED implements this rule, then compares the inexpensive density weight against an expensive gradient-based reference (Appendix~\ref{app:heed-variants}), and checks whether benchmark gains follow diagnostic prediction.

\section{HEED: Density-weighted residual alignment}
\label{sec:method}

The diagnostic provides the design rule for our method: Place more alignment weight on the positions that uniform distillation tends to lose. We turn this rule into a loss in two steps. First, we move the alignment target from per-layer block outputs to the residual stream, before adding any density weighting. This intermediate condition isolates whether the residual stream is a better alignment target; this yields C3: RSA. Second, we maintain the same residual-stream target but replace the uniform per-position weight with a density weight; this yields C4: HEED.

\begin{figure}[t]
\centering
\includegraphics[width=\linewidth]{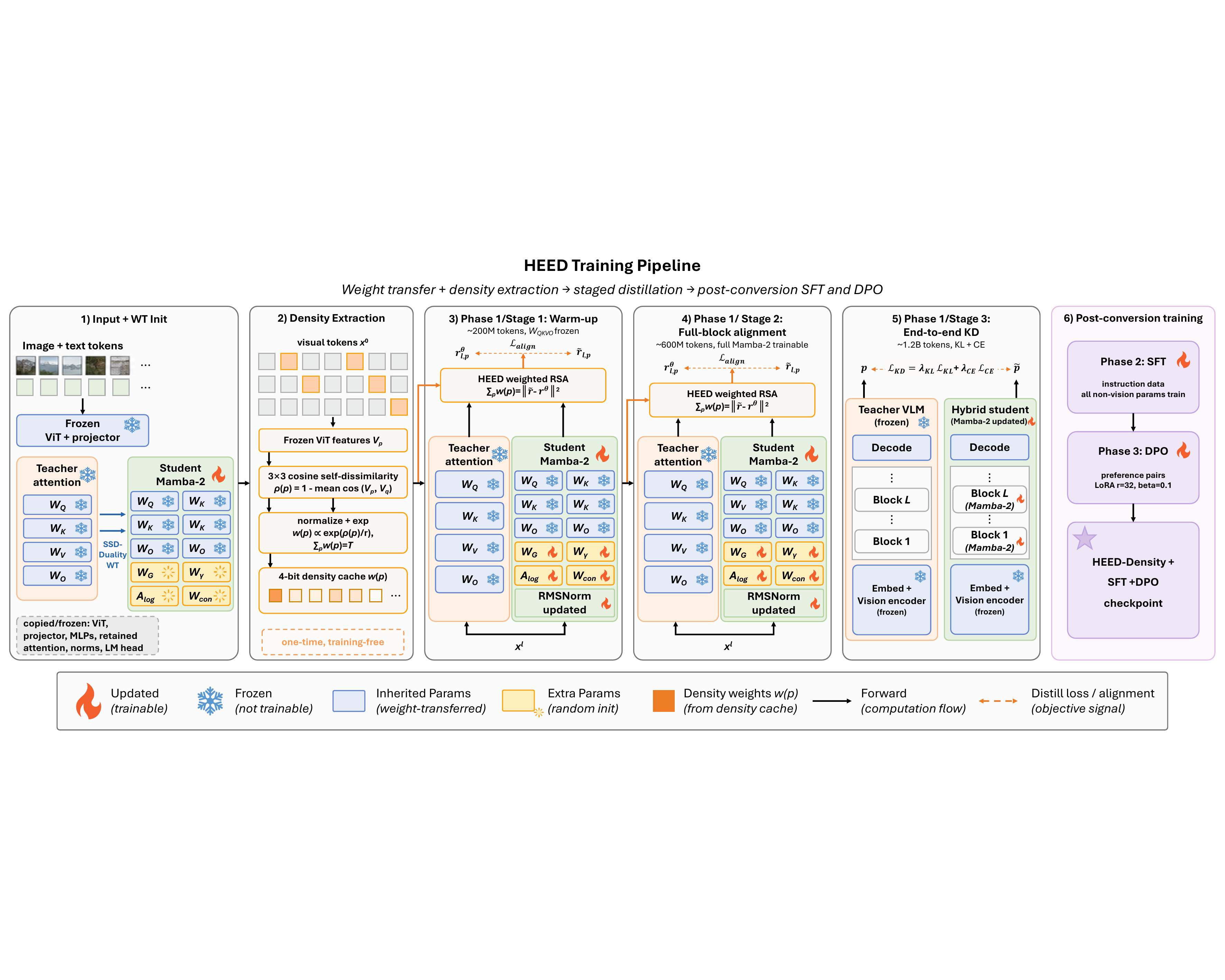}
\vspace*{-3mm}
\caption{\textbf{HEED training pipeline.} 
\textit{Left:} one-time weight initialization transfers teacher attention weights into the Student Mamba-2 blocks via SSD and the density cache $w(p)$ is precomputed from frozen ViT features (Eq.~\ref{eq:density}, Eq.~\ref{eq:density-weight}). 
\textit{Middle:} Stages~1/2 align teacher and student residual streams $r^{\theta}_{\ell,p}, r^{\tilde{\theta}}_{\ell,p}$ at every replaced layer with the density-weighted MSE $\mathcal{L}_{\text{align}}$; HEED differs from uniform residual alignment (C3) only in the per-position weight $w(p)$. 
\textit{Right:} Stage~3 end-to-end KD is shared across conditions C1-C4. All architecture, optimizer, and post-training settings are held identical across conditions; only the Stage~1/2 alignment loss differs. The density cache adds no inference-time cost.}
\label{fig:pipeline}
\vspace*{-3mm}
\label{fig:arch}
\end{figure}

The first step is useful because the residual stream is what later layers receive. Matching it asks each hybrid block to preserve the information passed forward through the network. The second step is needed because C3 is still uniform-weighted, which treats a blank background patch and a receipt digit as equally important alignment targets. HEED changes only this loss weighting and keeps the hybrid student architecture unchanged. It assigns each position a weight $w(p)$ from a one-time density cache computed from frozen ViT features. Once trained, the student is frozen at inference time. The full distillation pipeline of HEED is shown in Fig.~\ref{fig:arch}.

\subsection{Distillation setup}
\label{sec:method-prelim}

The setup is fixed across the main comparisons. Let $\teacher$ be the pretrained VLM and $\student$ the hybrid student. We replace 75\% of the decoder attention blocks with matched Mamba-2 blocks, resulting in the uniform 3:1 Mamba-2/attention hybrid used in the main experiments. The vision encoder, projector, root mean square normalization (RMSNorm), multilayer perceptrons (MLPs), retained-attention layers, language model head, and embeddings are copied from the teacher and frozen. Each new Mamba-2 block inherits $W_Q, W_K, W_V, W_O$ from the teacher attention block at the same layer via structured state space duality (SSD) weight transfer (WT) as initialization~\cite{dao2024mamba2,mmmamba2025}. The remaining Mamba-2 parameters are randomly initialized.

\paragraph{Three-stage distillation schedule.}
HEED includes three stages on a fixed token budget (10\% / 30\% / 60\%), following the staged structure of mmMamba~\cite{mmmamba2025}:
\begin{itemize}[leftmargin=1.4em,topsep=2pt,itemsep=1pt]
\item \emph{Stage 1 (warm-up)}: $W_{Q,K,V,O}$ frozen at their WT values, and only $W_G, W_\gamma, A$ and $W_{\text{conv}}$ train. Loss is the per-layer alignment term $\Lalign$.
\item \emph{Stage 2 (full block)}: full Mamba-2 blocks (including $W_{Q,K,V,O}$) train. Same $\Lalign$.
\item \emph{Stage 3 (end-to-end)}: full model trains end-to-end and the loss switches to standard logit KD:
\begin{equation}
\label{eq:kd}
\Lkd = \lambda_{\text{KL}}\Lkl(\teacher\,\|\,\student) + \lambda_{\text{CE}}\Lce(\student;\mathcal{D}).
\end{equation}
with $\lambda_{\text{KL}}{=}1.0$ and $\lambda_{\text{CE}}{=}0.1$. 
\end{itemize}
The experimental conditions described in
Sect.~\ref{sec:exp-setup} differ only in the Stage~1/2 alignment loss $\Lalign$ (or whether Stages~1/2 are run). Stage~3 uses the same KD objective and optimizer settings across conditions.

\subsection{Density as a lightweight importance signal}
\label{sec:method-density}

HEED needs a lightweight number that indicates which positions should receive more alignment weight. The diagnostic suggests a simple choice --- local feature distinctiveness. Smooth regions, such as sky, wall, skin, or blank page margins, look similar to nearby patches, whereas text characters, chart marks, edges, and object boundaries look different from nearby patches.

For each visual patch $p$, we compare its ViT feature $v_p$ with features in its $3{\times}3$ neighborhood $\mathcal{N}(p)$ as follows:
\begin{equation}
\label{eq:density}
\rho(p) = 1 - \frac{1}{|\mathcal{N}(p)|}\sum_{q\in\mathcal{N}(p)}
\frac{v_p^{\top}v_q}{\|v_p\|\|v_q\|}.
\end{equation}
A large $\rho(p)$ indicates that a patch is locally distinctive, whereas a small $\rho(p)$ indicates that it resembles its neighbors. The proxy uses two fixed design choices: a $3{\times}3$ neighborhood and cosine similarity. The neighborhood is the smallest nontrivial spatial window around a patch and cosine similarity provides a standard scale-invariant measure for ViT features. These choices are not tuned on the benchmark; the same defaults are used throughout. We then normalize density within each image to obtain $\tilde\rho(p)\in[0,1]$ and convert it into a loss weight:
\begin{equation}
\label{eq:density-weight}
w(p) \propto \exp\bigl(\tilde\rho(p)/\tau\bigr),\qquad \sum_p w(p)=T .
\end{equation}
Here, $T$ is the number of positions in the aligned sequence and $\tau$ is the temperature: smaller $\tau$ makes the weights concentrate more strongly on high-density positions. We use $\tau{=}0.5$ throughout, with no threshold, clipping, or smoothing. The normalization keeps the average loss scale comparable to RSA and only the allocation across positions changes. Appendix~\ref{app:method-recap}-\ref{app:theory} show the derivation of the gradient-based reference weight that density approximates.

\subsection{From uniform residual alignment to HEED}
\label{sec:method-loss}

We now define the C3 and C4 loss. Both extract the teacher and student residual streams after each replaced attention layer. C3 assigns every position the same weight:
\begin{equation}
\label{eq:rsa}
\Lrsa =
\frac{1}{|\mathcal{S}^*|\,T}
\sum_{\ell\in\mathcal{S}^*}\sum_{p=1}^{T}
\bigl\|\rconv_{\ell,p}-\rorig_{\ell,p}\bigr\|_2^2 .
\end{equation}
C4 maintains the same residual-stream target and changes only the per-position weight:
\begin{equation}
\label{eq:heed}
\Lheed =
\frac{1}{|\mathcal{S}^*|\,T}
\sum_{\ell\in\mathcal{S}^*}\sum_{p=1}^{T}
w(p)\,\bigl\|\rconv_{\ell,p}-\rorig_{\ell,p}\bigr\|_2^2 .
\end{equation}

The density weight $w(p)$ in Eq.~\ref{eq:density-weight} is computed from cached ViT density for visual residual positions. For non-visual positions, where no ViT patch density is available, we use a constant text-side density $\rho_{\mathrm{text}}=\beta \bar{\rho}_{\mathrm{visual}}$ with $\beta=2$, followed by the same exponential mapping and per-image normalization $\sum_p w(p)=T$. This keeps the total alignment budget fixed while changing how that budget is allocated across positions. The weight $w(p)$ depends on the input content but not on the layer index $\ell$, so the same cached/derived vector reweights every layer in $\mathcal{S}^*$. We set $\Lalign{=}\Lheed$ in Stages~1 and~2. Stage~3 then uses the KD loss $\Lkd$.

\section{Experiments}
\label{sec:experiments}

The experiments address one question. If the architecture, data, and budget stay fixed, does changing the hidden-state alignment to density-weighted residual alignment recover the fine-grained information lost by standard hybrid distillation?

\subsection{Setup and conditions}
\label{sec:exp-setup}

\paragraph{Base.} Qwen3-VL-8B-Instruct~\cite{qwen3vl2025} is the teacher. The student is the 3:1 Mamba-2 hybrid 
(Sect.~\ref{sec:method-prelim}).

\paragraph{Data.} Distillation uses a total of $\sim$1.34M samples ($\approx$ 2B tokens), blended following the InternVL family~\cite{chen2024internvl25}. Full data and optimizer details are presented in Appendix~\ref{app:recipes}. 

\paragraph{Evaluation.} We group benchmarks by what the diagnostic predicts should change, using lmms-eval~\cite{zhang2024lmmsevalrealitycheckevaluation}. \emph{Fine-grained perception (answering requires reading local symbols):} OCRBench~v2~\cite{ocrbenchv2}, DocVQA~\cite{docvqa}, ChartQA~\cite{chartqa}, TextVQA~\cite{textvqa}, InfoVQA~\cite{infovqa}, AI2D~\cite{ai2d}. \emph{Reasoning:} MMMU-Pro~\cite{mmmu_pro}, MMStar~\cite{mmstar}, MMBench~\cite{mmbench}, MathVista~\cite{mathvista}. 

\paragraph{Conditions.} Every condition uses the same backbone, data, hyperparameters, and WT initialization. The only differences are whether Stages~1/2 are used and which alignment loss $\Lalign$ they use. Stage~3 uses the same end-to-end logit KD for all four conditions, with $\lambda_{\text{KL}}{=}1.0$ and $\lambda_{\text{CE}}{=}0.1$~\cite{wang2024mambainllama}. C1 and C2 anchor the ladder to prior recipes. C2 is the recent prior VLM-specific hybrid-distillation baseline in our comparison. Appendix~\ref{app:priorwork} and Tab.~\ref{tab:priorwork-method} present the mapping to prior methods and the comparison between them.
\begin{itemize}[leftmargin=1.4em,topsep=2pt,itemsep=1pt]
\item \emph{C1 KD} (Mamba-in-Llama-style~\cite{wang2024mambainllama}): skip Stages~1/2; run Stage~3 on the full 2B-token budget.
\item \emph{C2 HSA} (mmMamba-style baseline~\cite{mmmamba2025}): Stages~1/2 use per-layer attention-output mean-squared error (MSE) ($\sum_\ell\|\hat O^\ell\!-\!O^\ell\|_2^2$), then Stage~3.
\item \emph{C3 RSA}: Stages~1/2 use uniform residual-stream MSE, then Stage~3. This is our internal control for the alignment target, which reads the residual stream instead of per-layer block outputs.
\item \emph{C4 HEED} (ours): same as C3 but Stages~1/2 use density-weighted residual MSE.
\end{itemize}
This ladder separates the effects we care about. C1$\to$C2 tests whether staged layer-wise alignment helps. C2$\to$C3 tests whether the residual stream is a better alignment target than per-layer block outputs. C3$\to$C4 tests density weighting itself.

\subsection{Main result: Uniform vs.\ density-weighted residual alignment}
\label{sec:exp-main}

Tab.~\ref{tab:main} reports scores after KD and before SFT+DPO. The main result is direct: HEED (C4) recovers much of the fine-grained gap left by normal end-to-end KD (C1). Compared with C1, C4 improves performance by 8.7 points on OCRBench~v2 by and 5.13 points on the 10-benchmark average.

In the controlled C3$\to$C4 contrast, density weighting adds 4.7 points on OCRBench~v2, 3.50 on the fine-grained benchmarks average, and 2.24 on the 10-benchmark average. The improvement is concentrated where the diagnostic predicts it should be concentrated: OCR, document, chart, and text-heavy benchmarks improve, whereas broad reasoning benchmarks remain stable (reasoning average 70.90 $\to$ 71.25). The fine-grained gains are not obtained by degrading reasoning.

The stepwise comparisons in Tab.~\ref{tab:main} lead to the same conclusion. Hidden-state matching helps (C1$\to$C2: +2.20 average) and moving the alignment target to the residual stream yields a smaller gain (C2$\to$C3: +0.69 average). Density weighting (C3$\to$C4) is the largest single step.

\subsection{Robustness in brief}
\label{sec:exp-robustness}
\label{sec:exp-cross-setting}

The controlled C3$\to$C4 gain is not tied to one run. A lower-cost 500M-token, three-seed replication preserves the controlled C3$\to$C4 gain at +4.8 $\pm$ 0.42 on OCRBench~v2 (Tab.~\ref{tab:seed-variance}, Appendix~\ref{app:seed-variance}), suggesting the effect is not a single-run artifact. The same controlled comparison also transfers across model choices: It yields +3.2 on InternVL-3.5~\cite{wang2025internvl3} (Tab.~\ref{tab:backbone}, Appendix~\ref{app:backbone-robustness}) and a 3:1 GDN hybrid trained with HEED reaches 59.9 on OCRBench~v2 with the same data (Tab.~\ref{tab:heed-variants}, Appendix~\ref{app:heed-variants}; no matched RSA control on GDN). The conclusion is that the density-weighting effect is stable across seeds and backbones. Appendix~\ref{app:signal-comparison} compares density with Sobel and OCR-localization weights at matched concentration.

\subsection{Practical endpoint: SFT+DPO}
\label{sec:exp-pipeline}

\begin{table}[t]
\centering\footnotesize
\caption{Benchmark scores after distillation and before SFT+DPO.}
\label{tab:main}
\setlength{\tabcolsep}{3.5pt}
\begin{tabular}{lrrrrr}
\toprule
            & \textbf{C0} & C1 & C2 & C3 & \textbf{C4} \\
            & Teacher & KD & HSA & RSA & \textbf{HEED} \\
\midrule
\multicolumn{6}{l}{\emph{Fine-grained perception}} \\
\midrule
OCRBench v2 & 63.8 & 50.5 & 53.1 & 54.5 & \textbf{59.2} \\
DocVQA      & 95.4 & 86.9 & 87.3 & 89.6 & \textbf{92.9} \\
ChartQA     & 88.1 & 77.9 & 81.0 & 81.8 & \textbf{83.4} \\
TextVQA     & 83.6 & 72.3 & 76.5 & 76.3 & \textbf{80.6} \\
InfoVQA     & 82.7 & 70.2 & 74.4 & 74.5 & \textbf{77.6} \\
AI2D        & 85.1 & 72.8 & 78.3 & 79.0 & \textbf{83.0} \\
\textbf{Fine-grained Avg. (6)} & \textbf{83.12} & 71.77 & 75.10 & 75.95 & \textbf{79.45} \\
\midrule
\multicolumn{6}{l}{\emph{Reasoning}} \\
\midrule
MMMU-Pro    & 55.2 & 53.9 & 54.4 & 55.0 & \textbf{55.0} \\
MMStar      & 70.2 & 68.9 & 69.1 & 69.8 & \textbf{70.1} \\
MMBench     & 84.0 & 82.1 & 82.9 & 83.3 & \textbf{83.7} \\
MathVista   & 76.0 & 74.9 & 75.4 & 75.5 & \textbf{76.2} \\
\textbf{Reasoning Avg. (4)} & \textbf{71.35} & 69.95 & 70.45 & 70.90 & \textbf{71.25} \\
\midrule
\textbf{Avg. (10 )} & \textbf{78.41} & 71.04 & 73.24 & 73.93 & \textbf{76.17} \\
\emph{$\Delta$ vs.\ teacher} & --     & -7.37 & -5.17 & -4.48 & \textbf{-2.24} \\
\midrule
\multicolumn{6}{l}{\emph{Controlled contrast: density weighting alone (same architecture, same data)}} \\
\midrule
\emph{$\Delta$ (C4 $-$ C3): OCRBench v2} & \multicolumn{5}{r}{\textbf{+4.70}} \\
\emph{$\Delta$ (C4 $-$ C3): fine-grained avg (6)} & \multicolumn{5}{r}{$\textbf{+3.50}$} \\
\emph{$\Delta$ (C4 $-$ C3): 10-benchmark avg} & \multicolumn{5}{r}{\textbf{+2.24}} \\
\midrule
\multicolumn{6}{l}{\emph{Overall gain over normal end-to-end KD}} \\
\midrule
\emph{$\Delta$ (C4 $-$ C1): OCRBench v2} & \multicolumn{5}{r}{\textbf{+8.70}} \\
\emph{$\Delta$ (C4 $-$ C1): 10-benchmark avg} & \multicolumn{5}{r}{\textbf{+5.13}} \\
\bottomrule
\end{tabular}
\end{table}

HEED is a distillation-stage intervention. Hence, the final question is whether the advantage survives normal post-training. We apply the same SFT+DPO recipe to every condition. The density-weighted student is the only row that reaches teacher-level performance on the 10-benchmark average under this evaluation suite (78.79 vs.\ 78.41, $\Delta=+$0.38), with per-benchmark differences in the range [-0.3, +1.2] around the teacher (Tab.~\ref{tab:full_pipeline_detailed}, Appendix~\ref{app:full-pipeline}). The selectivity predicted by the diagnostic also survives post-training: the fine-grained 6-benchmark average reaches 83.18 (teacher 83.12) while the reasoning 4-benchmark average reaches 72.20 (teacher 71.35). The uniform-weight pipelines remain 1.9-5.1 points below the teacher under the same post-training recipe (Tab.~\ref{tab:full_pipeline}).

\begin{table}[t]
\centering\footnotesize
\caption{Pipeline endpoints after identical SFT+DPO post-training (compact). Per-benchmark scores presented in Tab.~\ref{tab:full_pipeline_detailed}, Appendix~\ref{app:full-pipeline}.}
\label{tab:full_pipeline}
\setlength{\tabcolsep}{5pt}
\begin{tabular}{lrrrrr}
\toprule
& \textbf{C0} & C1+ & C2+ & C3+ & \textbf{C4+} \\
& Teacher & KD & HSA & RSA & \textbf{HEED} \\
\midrule
OCRBench v2                  & 63.8 & 57.6 & 59.5 & 61.2 & \textbf{63.9} \\
Fine-grained Avg.\ (6)       & 83.12 & 76.60 & 78.52 & 80.30 & \textbf{83.18} \\
Reasoning Avg.\ (4)          & 71.35 & 68.50 & 69.48 & 70.75 & \textbf{72.20} \\
\midrule
\textbf{Avg (10)} & \textbf{78.41} & 73.36 & 74.90 & 76.48 & \textbf{78.79} \\
\emph{$\Delta$ vs.\ teacher} & --     & -5.05 & -3.51 & -1.93 & \textbf{+0.38} \\
\bottomrule
\end{tabular}
\end{table}

The hybrid student maintains the inference profile of the 3:1 Mamba-2 architecture: 1.47$\times$ throughput at 4k context length, 2.84$\times$ at 32k, and 4.12$\times$ at 128k, with peak VRAM at 0.32$\times$ the teacher's at 128K. Time-to-first-token at 32k drops from 4.21\,s to 1.51\,s (2.8$\times$ faster). These numbers are derived from the hybrid architecture, which is identical across C1-C5. They support the practical endpoint: after the same post-training, HEED reaches teacher-level 10-benchmark average performance while maintaining the hybrid model's lower inference cost. Full per-context curves and per-condition breakdown are presented in Appendix~\ref{app:efficiency} (Tab.~\ref{tab:efficiency}, Fig.~\ref{fig:efficiency}).

\section{Limitations and conclusion}
\label{sec:conclusion}

Three limitations define the scope of the result:
\begin{itemize}[leftmargin=1.4em,topsep=2pt,itemsep=1pt]
\item \emph{Density is a proxy, not a universal saliency measure.} Patch self-dissimilarity is lightweight and works well in our setting but is not task sensitivity itself. Because the proxy is local distinctiveness, it naturally favors text, edges, and high-frequency visual structures. It may be less appropriate for tasks where the decisive evidence is globally defined, visually smooth, or adversarially patterned. In these regimes, density can misrank patches and the more expensive gradient-based C5 HEED-G variant presented in Appendix~\ref{app:heed-variants} is the intended fallback. We do not test shifted domains (medical, remote sensing, panoramic) or weights based on token entropy, teacher attention, or Grad-CAM. Density's success should be interpreted within the hybrid VLM conversion setting tested here.
\item \emph{Post-training is fixed.} We use the same SFT$+$DPO recipe for every condition to test whether the distillation-stage gain survives standard post-training. We do not tune a separate post-training recipe for each baseline. This controls the comparison but leaves open whether extra baseline-specific tuning could narrow some endpoint gaps.
\item \emph{Inference-time accelerators are not combined.} HEED is a training-time loss and does not remove tokens at inference time. Visual-token reduction methods, such as FastV, VisionZip, and VoCo~\cite{fastv2024,visionzip2024,voco2024}, work at inference time. Hence, they are complementary in principle and operate on the same per-patch importance signal that HEED uses but we do not test combined HEED + token-reduction systems here. All inference efficiency emanates from the hybrid architecture.
\end{itemize}
Additional scope discussion is presented in Appendix~\ref{app:extended-discussion}.

Hybrid VLM distillation does not fail uniformly. It can preserve broad reasoning while losing the small visual details needed for OCR and document understanding. HEED addresses this at the training-signal level: It gives more residual-alignment weight to dense visual positions, substantially reduces the main fine-grained gap, and does not add any parameters or inference-time cost. The broader lesson suggested by these results is that fixed-capacity hybrid sequence compressors may require distillation losses that reflect non-uniform token importance.

\begin{ack}
This work was supported by Qualcomm.
\end{ack}

\newpage
\bibliographystyle{plainnat}
\bibliography{refs}

\newpage
\appendix
\section{Extended results and additional experiments}
\label{app:extended-results}

This section presents supporting evidence for our main claims: translated prior-method context, HEED variants, matched alternative signals, efficiency, ablations, text-excluded diagnostics, seed checks, and residual failure modes.

\subsection{Translated prior-work comparison}
\label{app:priorwork}

The comparison presented here provides supporting context, not central evidence. Prior hybrid-distillation papers use different backbones, hybrid ratios, data, and evaluation harnesses: MaTVLM~\cite{matvlm2025} reports results on TinyLLaVA-Phi-2, mmMamba~\cite{mmmamba2025} on HoVLE-2.6B, and Mamba-in-Llama~\cite{wang2024mambainllama} on text-only Llama-3-8B. We, therefore, translate their loss and initialization choices to our shared setting: Qwen3-VL-8B-Instruct, a 3:1 Mamba-2 hybrid, and the same 2B-token budget.

\paragraph{How C1-C4 map onto prior methods.} Tab.~\ref{tab:priorwork-method} compares our four conditions with published hybrid-distillation recipes based on three design choices: initialization, layer-wise alignment used in Stages~1-2, and end-to-end Stage~3 loss. Two of our conditions are direct translations of published methods under our shared initialization and three-stage schedule: C1 KD follows the Mamba-in-Llama-style end-to-end KD recipe and C2 HSA follows mmMamba, the recent prior VLM-specific hybrid-distillation baseline in our comparison. C3 RSA and C4 HEED share C2's schedule but shift the alignment locus from per-layer attention output to the residual stream (C3 with uniform weighting, C4 with density weighting).

\begin{table}[h]
\centering\footnotesize
\caption{Recipe components in prior hybrid-distillation methods and our C1-C4 ladder. \emph{WT} = weight transfer initialization, \emph{MO} = matrix-orientation init, \emph{HSA} = hidden-state alignment, \emph{RSA} = residual-stream alignment, \emph{KD} = knowledge distillation, \emph{SFT}/\emph{DPO} = post-distillation fine-tuning. $^W$ marks density-weighted RSA (HEED).}
\label{tab:priorwork-method}
\setlength{\tabcolsep}{4pt}
\begin{tabular}{lccccccc}
\toprule
Method & WT & MO & HSA & RSA & KD  \\
\midrule
LoLCATs~\cite{zhang2024lolcats}            & \checkmark &            & \checkmark &            &            \\
MOHAWK / Llamba~\cite{bick2024mohawk}      &            & \checkmark & \checkmark &            & \checkmark \\
Mamba-in-Llama~\cite{wang2024mambainllama} & \checkmark &            &            &            & \checkmark  \\
MaTVLM~\cite{matvlm2025}                   & \checkmark &            & \checkmark &            & \checkmark  \\
mmMamba~\cite{mmmamba2025}                 & \checkmark &            & \checkmark &            & \checkmark \\
\midrule
\textbf{C1: KD}             & \checkmark &            &            &            & \checkmark \\
\textbf{C2: HSA}            & \checkmark &            & \checkmark &            & \checkmark  \\
\textbf{C3: RSA}            & \checkmark &            &            & \checkmark & \checkmark  \\
\textbf{C4: HEED (ours)}    & \checkmark &            &            & $\checkmark^{W}$ & \checkmark \\
\bottomrule
\end{tabular}
\end{table}

The controlled claim is the internal C3 RSA $\to$ C4 HEED contrast in Tab.~\ref{tab:main}. The mapping above only explains how C1-C4 relate to prior recipes.

\subsection{Relation to importance-aware and saliency-aware distillation}
\label{app:importance-aware}

Giving more informative tokens larger loss weights is a broad idea, and local self-dissimilarity is a classical descriptor~\cite{shechtman2007selfsim}. Existing importance-aware methods mainly reweight training examples, logits, task losses, or attention maps, and saliency-aware methods derive the weights from class-activation maps, such as Grad-CAM~\cite{selvaraju2017gradcam}. HEED addresses a narrower conversion problem: Which residual-stream positions should be preserved when full-attention blocks are replaced by a fixed-capacity hybrid mixer? It assigns position weights inside residual alignment, uses frozen input features, needs no labels or teacher backward pass, and leaves the inference path unchanged. Tab.~\ref{tab:importance-aware} summarizes the differences.

\begin{table}[h]
\centering\footnotesize
\caption{HEED compared with typical importance- and saliency-aware distillation.}
\label{tab:importance-aware}
\setlength{\tabcolsep}{4pt}
\begin{tabular}{p{0.2\linewidth}p{0.36\linewidth}p{0.36\linewidth}}
\toprule
Aspect & Typical importance/saliency distillation & HEED \\
\midrule
Main goal & Emphasize informative examples, outputs, or tokens & Repair selective position-level drift caused by attention replacement \\
Alignment target & Logits, task loss, features, or attention maps & Teacher residual stream at replaced layers \\
Importance source & Method-dependent; often teacher-, label-, or gradient-derived & Frozen input-side self-dissimilarity \\
Extra teacher backward pass & Required by gradient-based variants & Not required \\
\bottomrule
\end{tabular}
\end{table}

Our contribution, therefore, lies in the conversion diagnosis, sensitivity-weighted residual preservation, and the controlled finding that a training-free position signal recovers most of the gradient-weighted benefit (Appendices~\ref{app:heed-variants} and~\ref{app:signal-comparison}).

\subsection{Per-benchmark pipeline endpoints: SFT+DPO}
\label{app:full-pipeline}

Tab.~\ref{tab:full_pipeline_detailed} expands the compact main-text Tab.~\ref{tab:full_pipeline} with per-benchmark scores after identical SFT+DPO post-training. The benchmark grouping (fine-grained perception vs.\ reasoning) follows Sect.~\ref{sec:exp-setup}.

\begin{table}[h]
\centering\footnotesize
\caption{Per-benchmark pipeline endpoints after identical SFT+DPO post-training. Compact summary in Tab.~\ref{tab:full_pipeline}.}
\label{tab:full_pipeline_detailed}
\setlength{\tabcolsep}{5pt}
\begin{tabular}{lrrrrr}
\toprule
& \textbf{C0} & C1+ & C2+ & C3+ & \textbf{C4+} \\
& Teacher & KD & HSA & RSA & \textbf{HEED} \\
\midrule
\multicolumn{6}{l}{\emph{Fine-grained perception}} \\
\midrule
OCRBench v2 & 63.8 & 57.6 & 59.5 & 61.2 & \textbf{63.9} \\
DocVQA      & 95.4 & 89.4 & 90.9 & 92.5 & \textbf{95.2} \\
ChartQA     & 88.1 & 81.9 & 84.2 & 85.5 & \textbf{88.6} \\
TextVQA     & 83.6 & 77.7 & 79.3 & 81.3 & \textbf{83.7} \\
InfoVQA     & 82.7 & 75.7 & 77.5 & 79.4 & \textbf{82.4} \\
AI2D        & 85.1 & 77.3 & 79.7 & 81.9 & \textbf{85.3} \\
\textbf{Fine-grained Avg.\ (6)} & \textbf{83.12} & 76.60 & 78.52 & 80.30 & \textbf{83.18} \\
\midrule
\multicolumn{6}{l}{\emph{Reasoning}} \\
\midrule
MMMU-Pro    & 55.2 & 52.4 & 53.3 & 54.5 & \textbf{55.9} \\
MMStar      & 70.2 & 67.1 & 68.4 & 69.7 & \textbf{71.1} \\
MMBench     & 84.0 & 81.1 & 82.1 & 83.4 & \textbf{84.6} \\
MathVista   & 76.0 & 73.4 & 74.1 & 75.4 & \textbf{77.2} \\
\textbf{Reasoning Avg.\ (4)}    & \textbf{71.35} & 68.50 & 69.48 & 70.75 & \textbf{72.20} \\
\midrule
\textbf{Avg (10 benchmarks)} & \textbf{78.41} & 73.36 & 74.90 & 76.48 & \textbf{78.79} \\
\emph{$\Delta$ vs.\ teacher} & -     & -5.05 & -3.51 & -1.93 & \textbf{+0.38} \\
\bottomrule
\end{tabular}
\end{table}

\subsection{HEED variants on a fixed backbone}
\label{app:heed-variants}

Tab.~\ref{tab:heed-variants} extends Tab.~\ref{tab:main} with two additional HEED variants. Each variant changes one factor while keeping the rest of the recipe fixed: WT initialization, three-stage schedule, and 2B-token budget (Appendix~\ref{app:recipes-distill}). All three trainable-condition columns use the same Qwen3-VL-8B-Instruct teacher. C5 HEED-G is a diagnostic reference. It measures how much is lost when the expensive gradient-sensitivity weight is replaced by the lightweight density proxy, which is not a practical recommendation.

\begin{itemize}[leftmargin=1.4em,topsep=2pt,itemsep=1pt]
\item \textbf{C4 HEED} (reference, density-weighted): the canonical variant from Tab.~\ref{tab:main} on a 3:1 Mamba-2 hybrid student. Reproduced for ease of reading.
\item \textbf{C5 HEED-G} (gradient-weighted, diagnostic reference): replaces the visual-density proxy with the per-sample empirical-Fisher diagonal:
\[
w_{\text{grad}}^{(x)}(p) \;\propto\; \sum_{\ell\in\mathcal{S}^*}\bigl\|\nabla_{r_{\ell,p}}R_x(\rorig)\bigr\|_2^2
\]
(Eq.~\ref{eq:fisher-diag} in Appendix~\ref{app:method-recap},  derivation in Appendix~\ref{app:theory}). Same teacher and Mamba-2 mixer as C4. Principled but requires one cached teacher backward pass per sample: it raises 1.73$\times$ distillation compute of C1 KD (1.58$\times$ C4 HEED) for a 0.6 OCRBench~v2 points over C4 before post-training and only 0.3 points after SFT+DPO.
\item \textbf{C6 HEED-GDN}: keeps density weighting and the Qwen3-VL-8B as the teacher baseline but swaps the linear-time mixer from Mamba-2 to GDN at the same 3:1 ratio. Tests whether the gain is mixer-specific. We did not train a matched RSA control on GDN. Hence, C6 shows that HEED works with a second mixer but does not measure a controlled gain on GDN.
\end{itemize}

The cross-teacher robustness check (HEED on InternVL-3.5) is given in a separate table. Downstream sections (Appendix~\ref{app:efficiency}, Appendix~\ref{app:ablations}) reference C5 and C6 from this table for training-cost and per-factor analysis.

\begin{table}[h]
\centering\small
\caption{Per-benchmark comparison of HEED variants on Qwen3-VL-8B-Instruct, 2B-token distillation budget. The C4 HEED column is reproduced from Tab.~\ref{tab:main}. C5 HEED-G replaces the density proxy with a per-sample empirical-Fisher diagonal. C6 HEED-GDN replaces the Mamba-2 mixer with GDN at the same 3:1 ratio.}
\label{tab:heed-variants}
\setlength{\tabcolsep}{4pt}
\begin{tabular}{lcccc}
\toprule
            & \textbf{C0} & C4 HEED   & C5 HEED-G & C6 HEED-GDN \\
            & Teacher     & Mamba-2   & Mamba-2   & Gated DeltaNet \\
\midrule
\multicolumn{5}{l}{\emph{Fine-grained perception}} \\
\midrule
OCRBench v2 & 63.8 & 59.2 & 59.8 & 59.9 \\
DocVQA      & 95.4 & 92.9 & 93.0 & 92.4 \\
ChartQA     & 88.1 & 83.4 & 85.6 & 86.1 \\
TextVQA     & 83.6 & 80.6 & 80.6 & 80.2 \\
InfoVQA     & 82.7 & 77.6 & 79.5 & 78.9 \\
AI2D        & 85.1 & 83.0 & 81.8 & 82.4 \\
\midrule
\multicolumn{5}{l}{\emph{Reasoning}} \\
\midrule
MMMU-Pro    & 55.2 & 55.0 & 55.2 & 55.4 \\
MMStar      & 70.2 & 70.1 & 70.2 & 70.8 \\
MMBench     & 84.0 & 83.7 & 83.8 & 84.6 \\
MathVista   & 76.0 & 76.2 & 75.9 & 77.9 \\
\midrule
\textbf{Avg (10 benchmarks)} & \textbf{78.41} & 76.17 & 76.54 & 76.86 \\
\emph{$\Delta$ vs.\ teacher} & ---            & -2.24        & -1.87        & -1.55        \\
\bottomrule
\end{tabular}
\end{table}

The small C5 gain over C4 (+0.6 before post-training, +0.3 after SFT+DPO) at 1.58$\times$ the C4 distillation cost, together with the cross-layer Spearman $\bar\rho_S{=}0.73$ (Appendix~\ref{app:theory}) and the per-sample $\rho(p)$-$\|g_{\ell,p}\|_2^2$ rank correlation of 0.63 overall (0.71 in the upper-density tail), are jointly consistent with density being a low-cost proxy for the diagonal-Fisher weight on this evaluation suite. We treat C5 HEED-G as a reference rather than a practical method.

\subsection{Alternative importance signals at matched concentration}
\label{app:signal-comparison}

High-density patches often contain text and edges. Hence, density may only be a cheap correlate for them. We test this by replacing density with two other visual signals. \emph{Sobel gradient magnitude}, averaged within each patch, is a standard pure-edge signal. \emph{OCR localization} uses each patch's overlap with boxes from the same text detector as the one discussed in Appendix~\ref{app:text-control} and tests whether density mainly finds text. Both arms follow the 500M-token C3 RSA protocol of Tab.~\ref{tab:seed-variance} and change only the per-patch visual signal. The architecture, data mixture, token budget, residual target, Stage~3 KD, text boost $\beta{=}2$, per-image normalization, exponential mapping with $\tau{=}0.5$, and mean weight stay fixed. The realized Gini coefficients of the weights are 0.296/0.293/0.289 for density/Sobel/OCR. Weight concentration, therefore, differs by at most 2.4\% and does not explain the ordering.

\begin{table}[h]
\centering\footnotesize
\caption{Alternative visual weight signals under the same 500M-token RSA protocol. C3 RSA and density show mean $\pm$ std over three seeds (Tab.~\ref{tab:seed-variance}); Sobel and OCR localization are single runs. \emph{Non-OCR Avg.} averages the five fine-grained benchmarks other than OCRBench~v2.}
\label{tab:signals-full}
\setlength{\tabcolsep}{3.5pt}
\begin{tabular}{lcccccc}
\toprule
Weight signal & Gini & OCRBench v2 & \makecell{Fine-grained\\Avg.\ (6)} & \makecell{Non-OCR\\Avg.\ (5)} & \makecell{Reasoning\\Avg.\ (4)} & Avg (10) \\
\midrule
Uniform (C3 RSA)        & 0     & 53.0 $\pm$ 0.6 & 73.43 $\pm$ 0.5 & 77.52 & 70.46 & 72.24 $\pm$ 0.35 \\
Sobel gradient          & 0.293 & 55.1 & 74.45 & 78.32 & 70.35 & 72.81 \\
OCR localization        & 0.289 & 55.9 & 74.76 & 78.52 & 70.30 & 72.98 \\
\textbf{Density (C4 HEED)} & 0.296 & \textbf{57.8 $\pm$ 0.4} & \textbf{75.87 $\pm$ 0.4} & \textbf{79.48} & \textbf{70.75} & \textbf{73.82 $\pm$ 0.29} \\
\bottomrule
\end{tabular}
\end{table}

\begin{table}[h]
\centering\footnotesize
\caption{Per-benchmark margin of density over OCR-localization weighting (500M tokens; density minus OCR, unrounded).}
\label{tab:signal-margins}
\setlength{\tabcolsep}{5pt}
\begin{tabular}{lccccccc}
\toprule
 & OCRBench v2 & DocVQA & ChartQA & TextVQA & InfoVQA & AI2D & Fine-grained Avg.\ (6) \\
\midrule
Density $-$ OCR & +1.87 & +0.87 & +0.85 & +0.58 & +1.30 & +1.19 & +1.11 \\
\bottomrule
\end{tabular}
\end{table}

Density leads OCR localization on all six fine-grained benchmarks (Tab.~\ref{tab:signal-margins}), including ChartQA and AI2D. Hence, the advantage extends beyond OCR-heavy benchmarks. Reasoning averages stay within 0.5 points across all four arms (Tab.~\ref{tab:signals-full}). The Sobel and OCR arms are each trained once. Hence, their ordering is directional rather than statistically resolved.

Empirical-Fisher weighting (C5 HEED-G, Appendix~\ref{app:heed-variants}) is the principled reference. At 2B tokens, it reaches 59.8 on OCRBench~v2 vs.\ 59.2 for density but costs 260 vs.\ 165 GPU-hours. Density's mean per-image Spearman correlation with layer-summed gradient sensitivity is 0.63 overall and 0.71 in the upper-density tail. Density is, therefore, a strong training-free proxy with a favorable trade-off between accuracy and cost. We do not claim that it is the uniquely correct importance signal. We do not test token entropy, teacher attention, or Grad-CAM-style saliency as weighting signals. Teacher attention concentration appears only as a predictor in the diagnostic regression, where its unique contribution is small (Appendix~\ref{app:text-control}).

\subsection{Cross-backbone robustness}
\label{app:backbone-robustness}

Tab.~\ref{tab:backbone} extends the C0-C4 ladder of Tab.~\ref{tab:main} to three additional VLM teachers. Each teacher is distilled into a 3:1 Mamba-2 hybrid with the same WT initialization, three-stage schedule, and 2B-token budget (Appendix~\ref{app:recipes-distill}). We report \emph{OCRBench~v2}, the central diagnostic benchmark. The C4+ column shows the endpoint after the shared SFT+DPO pipeline (Phase~2 + Phase~3 in Appendix~\ref{app:recipes}).

\begin{table}[h]
\centering\small
\caption{Cross-backbone robustness on OCRBench~v2: C0-C4 distillation ladder and C4+SFT+DPO endpoint, on three teachers other than Qwen3-VL-8B-Instruct (results for which are presented in Tab.~\ref{tab:main} and Tab.~\ref{tab:full_pipeline}). All students are 3:1 Mamba-2 hybrids of the corresponding teacher. The C3$\to$C4 step is positive on every backbone.}
\label{tab:backbone}
\setlength{\tabcolsep}{4pt}
\begin{tabular}{lcccccc}
\toprule
Backbone (teacher)         & \textbf{C0} & C1 & C2  & C3  & \textbf{C4}   & \textbf{C4+} \\
                           & Teacher     & KD & HSA & RSA & \textbf{HEED} & \textbf{HEED+SFT+DPO} \\
\midrule
InternVL-3.5-8B            & 53.2 & 43.8 & 46.6 & 47.1 & \textbf{50.3} & \textbf{52.6} \\
MiniCPM-V 4.5 (8B)         & 58.8 & 48.9 & 50.7 & 52.5 & \textbf{55.2} & \textbf{59.1} \\
GLM-4.6V-Flash (9B)        & 62.3 & 51.1 & 54.3 & 55.0 & \textbf{58.7} & \textbf{61.2} \\
\bottomrule
\end{tabular}
\end{table}

The controlled C3$\to$C4 step is positive on every backbone: +3.2 on InternVL-3.5~\cite{wang2025internvl3}, +2.7 on MiniCPM-V~4.5~\cite{yu2025minicpm}, +3.7 on GLM-4.6V-Flash~\cite{vteam2025glm45vglm41vthinkingversatilemultimodal}, all measured on OCRBench~v2. 

After the shared SFT+DPO pipeline, the C4+ row reaches the teacher's OCRBench~v2 performance to within $\pm $1 point on each backbone (InternVL: -0.6, MiniCPM-V: +0.3, GLM: -0.9). The density-weighting effect is, therefore, not specific to Qwen3-VL among the tested 8-9B teachers.

\subsection{Efficiency}
\label{app:efficiency}

\begin{figure}[h]
\centering
\includegraphics[width=\linewidth]{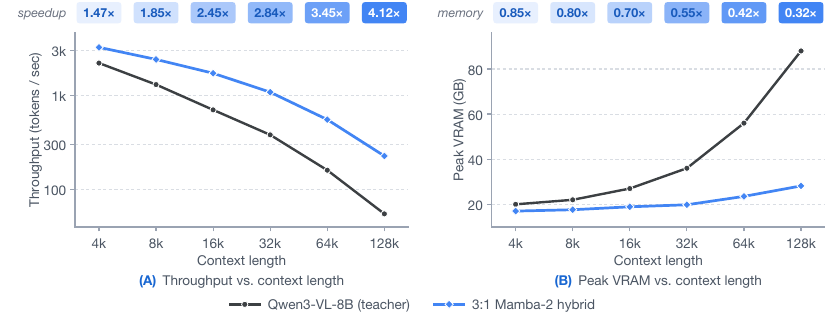}
\vspace*{-2mm}
\caption{\textbf{Inference efficiency vs.\ context length, measured with vLLM on a single H100 for the teacher and the 3:1 Mamba-2 hybrid student.} The hybrid architecture's throughput advantage grows with context (left) and its peak VRAM is roughly flat where the teacher's grows quadratically (right). HEED leaves the inference path unchanged. All hybrid students (C1-C5) share these curves.}
\vspace*{-3mm}
\label{fig:efficiency}
\end{figure}

Efficiency has two parts. At inference time, C1-C5 share the same hybrid backbone. Therefore, they share the same speed and memory profile relative to the teacher: 4.12$\times$ throughput and 0.32$\times$ peak VRAM at 128k, and $\sim$2.8$\times$ lower time-to-first token (TTFT) at 32k (Tab.~\ref{tab:efficiency} and Fig.~\ref{fig:efficiency}). At training time, C4 HEED stays close to C1 KD, whereas C5 HEED-G is slower because it requires a cached teacher backward pass. The inference gains, therefore, are derived from the 3:1 hybrid architecture. HEED recovers accuracy without changing the inference path: It adds no inference-time parameters or computation and costs 165 GPU-hours vs.\ 160 for C3 RSA.

\begin{table}[h]
\centering\footnotesize
\caption{Inference and training efficiency. \emph{Inference:} all
Mamba-2 hybrids match because the architecture is identical across C1-C5. \emph{Training:} C5 HEED-G inflates distillation compute by $\sim$1.7$\times$. Inference numbers measured with vLLM on a single H100.}
\label{tab:efficiency}
\setlength{\tabcolsep}{5pt}
\begin{tabular}{lccccc}
\toprule
 & \multicolumn{3}{c}{Inference} & \multicolumn{2}{c}{Training} \\
\cmidrule(lr){2-4}\cmidrule(lr){5-6}
Condition & \makecell{Throughput\\@128k (rel.)} & \makecell{Peak VRAM\\@128k (rel.)} & \makecell{TTFT\\@32k (s)} & \makecell{GPU-hr\\/ 2B tok} & rel.\ to KD \\
\midrule
Qwen3-VL-8B-Instruct (C0 teacher) & 1.00$\times$ & 1.00$\times$ & 4.21 & -   & -     \\
C1: KD                & 4.12$\times$ & 0.32$\times$ & 1.51 & 150  & 1.00$\times$ \\
C2: HSA               & 4.12$\times$ & 0.32$\times$ & 1.51 & 158  & 1.05$\times$ \\
C3: RSA               & 4.12$\times$ & 0.32$\times$ & 1.51 & 160  & 1.07$\times$ \\
C4: HEED     & 4.12$\times$& 0.32$\times$ & 1.51 & 165  & 1.10$\times$ \\
C5: HEED-G            & 4.12$\times$& 0.32$\times$ & 1.51 & 260 & 1.73$\times$ \\
\bottomrule
\end{tabular}
\end{table}

C5 is a diagnostic reference, not a practical recommendation. It costs 58\% more GPU-hours over C4 HEED for post-pipeline gains of only 0.3 on OCRBench~v2 and 0.25 on an average, both within single-seed evaluation noise. For completeness, the C5 HEED-G post-training endpoint is OCRBench v2 64.2 and 10-benchmark average 78.94 ($+0.15$ over C4+, $+0.53$ over the teacher), within single-seed evaluation noise of C4+.
\subsection{Ablations}
\label{app:ablations}

We evaluate ablations on OCRBench~v2, the central diagnostic. All rows use the same 2B-token, single-seed protocol as Tab.~\ref{tab:main}. C0 (teacher) and C4 (HEED, default) are reproduced from Tab.~\ref{tab:main} as anchors. As shown in Tab.~\ref{tab:ablations}, the method is robust to nearby design choices, including the residual locus, density granularity, and layer dependence. 

\begin{table}[h]
\centering\footnotesize
\caption{Ablation variants on OCRBench~v2. Each row removes or modifies one component of HEED relative to the C4 default (3:1 Mamba-2 hybrid, density-weighted residual MSE in Stages~1/2). The 3:1 ratio is the standard hybrid choice in this regime and is held fixed.}
\label{tab:ablations}
\setlength{\tabcolsep}{6pt}
\begin{tabular}{lcl}
\toprule
Condition & OCRBench v2 & Note \\
\midrule
C0 (Teacher)        & 63.8 & --- \\
C4 (HEED, default)  & 59.2 & 3:1 Mamba-2 hybrid, density-weighted residual MSE \\
\midrule
$-$Residual         & 56.7 & shift locus from residual stream to per-layer block-output MSE (C2 $+$ density) \\
$-$Density          & 56.6 & remove per-patch density, text/visual modality boost only \\
$+$LD               & 58.6 & layer-dependent weights instead of one shared weight \\
$+$SL               & 58.1 & single-layer gradient reference instead of layer-summed \\
\midrule
\multicolumn{3}{l}{\emph{Local weighting checks, 500M-token protocol of Tab.~\ref{tab:seed-variance}}} \\
\midrule
C3 (RSA)              & 53.0 $\pm$ 0.6 & three seeds \\
C4 (HEED, default)                  & 57.8 $\pm$ 0.4 & $\tau{=}0.5$, $\beta{=}2$; three seeds \\
$\beta{=}1$                   & 57.1 & no text boost; single seed \\
$\tau{=}1.0$                  & 56.9 & higher temperature; single seed \\
\bottomrule
\end{tabular}
\end{table}

The pattern across rows is direct.
\emph{(i)} \textbf{The alignment target matters.} $-$Residual applies density weighting to per-layer block-output MSE (the HSA locus, i.e., C2 with density) instead of the residual stream and underperforms C4 by 2.5 points. The gain is, therefore, not just density-aware reweighting. It is density-aware \emph{residual-stream} alignment.
\emph{(ii)} \textbf{Patch-level weighting matters.} $-$Density uses only a text/visual modality boost, with no per-patch signal. It recovers only part of HEED's gain, showing that the useful predictor is per-patch self-dissimilarity rather than the coarser text-vs-visual split.
\emph{(iii)} \textbf{One shared position weight is stable.} $+$LD uses per-layer weights and gives no reliable mean gain, whereas $+$SL uses a single-layer gradient reference and sits below the layer-summed reference. We, therefore, use one layer-summed weight per position.

These ablations support the final recipe: One per-position residual-alignment weight, normalized per sample, approximating the layer-summed gradient sensitivity with patch self-dissimilarity. In the lower block, removing the text boost ($\beta{=}1$) results in 57.1 and raising the temperature to $\tau{=}1.0$ results in 56.9. Both retain most of the default's gain over C3 RSA (57.8 $\pm$ 0.4 vs.\ 53.0 $\pm$ 0.6). These are single-seed checks around the defaults. The defaults, $\tau{=}0.5$, $3{\times}3$ neighborhood, and text boost $\beta{=}2$, are used throughout.

\subsection{Drift-variance decomposition: Bootstrap procedure}
\label{app:density-decomposition}

Next, we present the bootstrap procedure used to obtain the 95\% confidence interval for the density semi-partial $R^2$ reported in Sect.~\ref{sec:diag-density}. Uncertainty is estimated by image-level resampling of the diagnostic set: each 1{,}000 bootstrap resamples the 1{,}000 diagnostic images with replacement, refits the token-level linear regression on the resampled data, and recomputes the joint and semi-partial $R^2$ values. The resulting 95\% confidence interval for the density semi-partial $R^2$ is [0.27,0.33] and density is the largest single predictor in every resample ($p{<}10^{-3}$). The unexplained $\sim$0.47 of variance is expected: drift also depends on measurement noise, within-density variation, interactions among factors, and factors not measured here. Appendix~\ref{app:text-control} discusses the addition of OCR overlap and Sobel gradient magnitude to this regression as direct text and edge controls.

\subsection{Text-excluded diagnostic and extended regression}
\label{app:text-control}

Text is a common high-density case: The fine-grained benchmarks require reading it. Text-bearing patches alone, therefore, cannot establish the diagnosis in Sect.~\ref{sec:diag-density}. We add two direct controls on the same 1{,}000-sample diagnostic set.

\paragraph{Detector-negative patches.} We use DeepSeek-OCR~\cite{wei2025deepseekocr} as the text detector (0.96 recall, 0.99 precision) and repeat the top- vs.\ bottom-density comparison on patches with no detected text. Because recall is below 1, we call these patches \emph{detector-negative} rather than non-text. Their bottom- vs.\ top-density residual drift is 0.074 vs.\ 0.215 (2.91$\times$; 10{,}000-replicate image-bootstrap 95\% CI [2.71, 3.07]) and their teacher-masking answer sensitivity is 0.047 vs.\ 0.101 (2.15$\times$; CI [1.74, 2.65]). Both intervals exclude 1.

\paragraph{Manual false-negative audit.} A blind manual audit of 100 top-density and 100 bottom-density detector-negative patches found missed text in four patches in each group. This sample is too small to prove equal miss rates. Removing every confirmed missed-text patch from the audited sample still leaves drift and sensitivity ratios of 2.48$\times$ and 2.03$\times$.

\paragraph{Six-predictor regression.} We refit the drift regression of Sect.~\ref{sec:diag-density} with two added predictors, fixed before fitting: patch overlap with detected text (OCR overlap) and Sobel gradient magnitude. Tab.~\ref{tab:six-predictor} reports the semi-partial $R^2$ values. The full-model $R^2$ rises only from 0.53 with four predictors to 0.54. Density's unique contribution falls from 0.30 to 0.16 because it shares variance with the two new predictors, but density remains the largest measured predictor.

\begin{table}[h]
\centering\footnotesize
\caption{Semi-partial $R^2$ of residual drift in the six-predictor regression (full-model $R^2{=}0.54$). The CI is from an image-level bootstrap.}
\label{tab:six-predictor}
\setlength{\tabcolsep}{6pt}
\begin{tabular}{lc}
\toprule
Predictor & Semi-partial $R^2$ \\
\midrule
Density           & \textbf{0.16} [0.10, 0.26] \\
OCR overlap       & 0.07 \\
Token type        & 0.06 \\
Sobel gradient magnitude & 0.04 \\
Layer depth       & 0.03 \\
Teacher attention concentration & 0.02 \\
\bottomrule
\end{tabular}
\end{table}

The position-level pattern is, therefore, not explained by text location alone.

\paragraph{Why might dense positions drift?} We offer an architectural hypothesis. Full attention keeps direct token-to-token interactions, whereas a recurrent or state-space mixer must compress a growing prefix into a fixed-size state. This bottleneck may smooth sparse, locally distinct information more readily than redundant background. If so, drift should concentrate on dense non-text content as well as on OCR patches. The detector-negative ratios and density's leading semi-partial $R^2$ after text and edge controls fit this prediction but do not prove the mechanism.

\subsection{Density-targeted upweighting control}
\label{app:density-control}

The main diagnostic shows that the measured residual drift and teacher-masking answer sensitivity concentrate on high-density positions. We also check that the resulting OCRBench~v2 gain is not simply explained by giving extra loss weight to any subset of positions.

We re-run hybrid distillation with a binary-mask alignment loss. For each value of $k \in \{0,10,25,50\}$, we give a 5$\times$ weight to $k\%$ of positions and leave the rest unchanged. We compare two ways to choose the weighted positions. The density-targeted rule chooses the top-$k\%$ by $\rho(p)$. The random control chooses the same number of positions uniformly at random, with three random seeds for each $k$. The $k=$0 point is the C1 baseline. The continuous HEED weighting used in the main method is shown in Fig.~\ref{fig:diagnostic} as a reference endpoint: it is not a binary mask.

The two curves start from the same 50.5 OCRBench~v2 baseline. With density-targeted weighting, the score rises to 54.0, 58.0, and 58.7 at $k=$10, 25, 50, respectively. With random weighting, it reaches only 51.5, 52.8, and 54.0, respectively. Thus, density-targeted weighting is better by 2.5, \textbf{5.2}, and 4.7 points at the same mask sizes. Random-mask scores at each $k$ are averaged across three seeds (std $\le 0.6$ on OCRBench v2 for all $k$); the 2.5--5.2 point gap to density-targeted weighting is well outside this noise. This is a controlled intervention: Only the choice of which positions are upweighted changes between the two arms. This also supports the main interpretation: Selecting high-density positions matters more than selecting the same number of arbitrary positions.

\subsection{Seed variance: 500M-token three-seed replication}
\label{app:seed-variance}

To estimate seed-level noise under a comparable protocol, we run a lower-cost 500M-token, three-seed replication of C3 RSA $\to$ C4 HEED. All other recipe choices are fixed: Architecture, data mixture, optimizer, and evaluation harness match Tab.~\ref{tab:main}. Tab.~\ref{tab:seed-variance} reports per-seed means with standard deviation across the three seeds; the delta row reports the mean $\pm$ standard error (SE) of the paired difference. The +4.8 $\pm$ 0.42 OCRBench~v2 figure cited in Sect.~\ref{sec:exp-robustness} comes from this run.

\begin{table}[h]
\centering\footnotesize
\caption{500M-token, three-seed replication of the central C3 RSA $\to$ C4 HEED contrast. The C0 teacher row is reproduced from Tab.~\ref{tab:main} as an anchor (the teacher is deterministic; no seed variance). C3/C4 cells show mean $\pm$ std across $n{=}3$ seeds. The delta row shows the mean $\pm$ standard error of the paired differences, where $\Delta_i=\mathrm{C4}_i-\mathrm{C3}_i$ is computed for each seed and $\mathrm{SE}_\Delta=\mathrm{std}(\{\Delta_i\}_{i=1}^3)/\sqrt{3}$.}
\label{tab:seed-variance}
\setlength{\tabcolsep}{6pt}
\begin{tabular}{lccc}
\toprule
Condition & OCRBench v2 & Fine-grained avg (6) & 10-benchmark avg \\
\midrule
\textbf{C0} Teacher                 & 63.8 & 83.12 & 78.41 \\
\midrule
C3 RSA                              & 53.0 $\pm$ 0.6 & 73.43 $\pm$ 0.5 & 72.24 $\pm$ 0.35 \\
C4 HEED                             & 57.8 $\pm$ 0.4 & 75.87 $\pm$ 0.4 & 73.82 $\pm$ 0.29 \\
\midrule
\textbf{C4 $-$ C3 (mean $\pm$ SE)}  & \textbf{$+$4.8 $\pm$ 0.42} & \textbf{$+$2.44 $\pm$ 0.37} & \textbf{$+$1.58 $\pm$ 0.26} \\
\bottomrule
\end{tabular}
\end{table}

The 500M-token OCRBench~v2 gap (+4.8) matches the 2B-token main-table gap (+4.7). The central conclusion is preserved at both budgets: Density-weighted residual alignment beats uniform residual alignment under matched architecture, data, and budget.

\subsection{Extended discussion}
\label{app:extended-discussion}

The scope of the claim is specific. Under standard hybrid VLM distillation, the measured residual drift and teacher-masking answer sensitivity concentrate on high-density visual positions. Density is the strongest tested predictor of that drift among the factors we measured. An empirical-Fisher argument motivates importance-weighted residual alignment. The lightweight density proxy recovers most of the gradient-weighted benefit on this evaluation suite without changing inference. We do not claim Pareto dominance over full attention, nor that patch self-dissimilarity is a universal saliency measure, nor that the diagnostic decomposition exhausts all causes of conversion drift. All reported inference efficiency comes from the hybrid architecture. The matched-signal comparison supports density as a strong training-free proxy, not as the uniquely correct importance signal (Appendix~\ref{app:signal-comparison}).

\paragraph{Why the C3$\to$C4 contrast is informative.}
C3 RSA and C4 HEED share the same backbone, data, token budget, alignment target (residual stream), Stage~3 KD, and trainable parameters, and the only changed factor is the per-position weight in the residual-alignment loss. We treat this contrast as the cleanest test in this work: It isolates loss shape from architecture, data, and target choice. Under uniform weights, sparse high-density positions contribute little to the average alignment gradient. HEED reallocates this fixed total alignment budget toward positions that the diagnostic identifies as both drifting more and mattering more for the teacher's answer.

\begin{figure}[h]
\centering
\includegraphics[width=\linewidth]{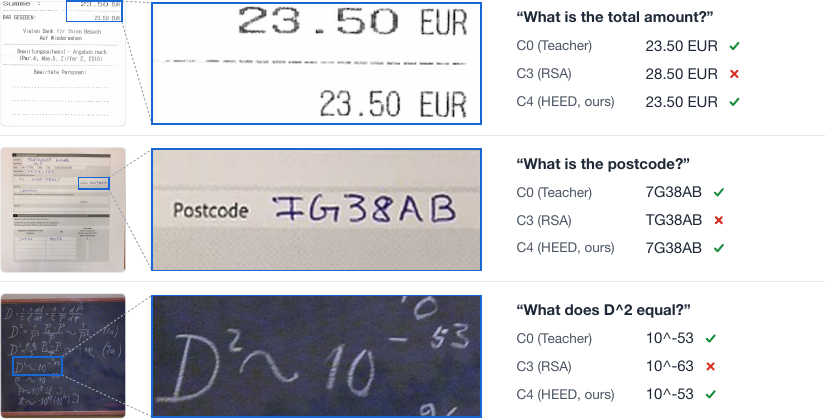}
\vspace*{-2mm}
\caption{\textbf{Concrete OCR failures.} C3 misreads a key character, while C4 matches the teacher or gold answer.}
\label{fig:failure-examples}
\end{figure}

\paragraph{Qualitative failure examples.}
Fig.~\ref{fig:failure-examples} shows the same pattern at the example level. These are not cases where the hybrid student misses the broad image content. C3 can usually identify that the image is a receipt, form, or written expression but a single local symbol is copied incorrectly and the final answer changes. C4 corrects these examples in the direction that the diagnostic predicts: The density-weighted loss assigns more alignment weight to the small distinctive regions where characters, digits, and marks live.

\paragraph{Where HEED helps less.}
HEED helps less in two regimes. First, reasoning-dominant inputs often have low or diffuse density. Hence, the selective signal is weak and gains over C3 RSA sit within noise. Second, density can misrank positions when visual distinctiveness and task importance diverge: adversarial textures can be locally distinctive without being useful, while some answer-bearing regions can be visually smooth. In these cases, C5 HEED-G is the expensive fallback because it measures gradient sensitivity directly. A hand-categorization of 50 OCRBench~v2 cases, where both C1 KD and C4 HEED fail, isolates three residual-error modes that account for $\ge$ 75\% of joint failures: \textbf{(F1)} multi-step reasoning over correctly extracted text. \textbf{(F2)} out-of-distribution scripts and decorative or handwritten fonts, where the ViT features are less distinctive. \textbf{(F3)} ultra-fine-grained marks below the ViT's 14$\times$14 patch resolution.

\paragraph{Broader impact.}
Lower-cost VLM inference can help resource-constrained uses, such as document accessibility and education, but the same efficiency applies to surveillance and high-volume document processing. HEED changes a training recipe rather than model access controls and deployments should inherit the same dual-use review expected for the base model.

\section{Methodological detail}
\label{app:methodology}

This section presents method details needed for scrutiny and reproduction: definitions, staged HEED pipeline, density computation, and empirical-Fisher motivation.

\subsection{HEED method definitions}
\label{app:method-recap}

Notation: $\teacher / \student$ represent teacher/student models, $\mathcal{S}$ is the set of attention layers replaced by Mamba-2 blocks, $\mathcal{S}^*=\{\ell+1:\ell\in\mathcal{S}\}$ represents the residual-alignment readout layers, $r^\theta_{\ell,p}/\rconv_{\ell,p}$ represent teacher/student residual streams, $g_{\ell,p}{=}\nabla_{r_{\ell,p}}R_x(\rorig)$ is the teacher-loss gradient, $\rho(p)$ is patch self-dissimilarity, and $w(p)$ is the normalized per-position alignment weight.

\paragraph{Base loss and HEED replacement.}
The uniform baseline uses CE, KL, and hidden-state or residual MSE with equal per-position weights. HEED keeps CE and KL unchanged but replaces the uniform alignment term with
\begin{equation}
\Lheed = \frac{1}{|\mathcal{S}^*|\,T}\sum_{\ell\in\mathcal{S}^*}\sum_{p=1}^{T} w(p)\,\bigl\|\rconv_{\ell,p}-\rorig_{\ell,p}\bigr\|_2^2 ,
\end{equation}
where $\Lheed$ is used as the alignment term $\Lalign$ in Stages~1 and~2 of the three-stage distillation schedule (Sect.~\ref{sec:method-prelim}). Stage~3 then runs the shared KD loss $\Lkd=\lambda_{\text{KL}}\Lkl+\lambda_{\text{CE}}\Lce$ end-to-end. Alignment reads residual streams after each replaced attention layer. $\Lheed$ replaces per-layer hidden-state matching as the locus of the alignment term, not as an additional term on top of it. Tab.~\ref{tab:condition-loss} summarizes the per-condition Stages~1/2 alignment loss for all four ladder rows.

\begin{table}[h]
\centering\footnotesize
\caption{Per-condition Stages~1/2 alignment loss $\Lalign$. Stage~3 is identical across conditions ($\Lkd$).}
\label{tab:condition-loss}
\setlength{\tabcolsep}{6pt}
\begin{tabular}{lll}
\toprule
Condition & $\Lalign$ in Stages~1/2 & Notes \\
\midrule
C1 (KD)         & --- (Stages~1/2 skipped)    & WT init $\to$ Stage~3 only, Mamba-in-Llama-style \\
C2 (HSA)        & per-layer attn-output MSE   & $\sum_{\ell}\|\hat O^\ell\!-\!O^\ell\|_2^2$, mmMamba-style \\
C3 (RSA)        & uniform residual-stream MSE & $\sum_{\ell\in\mathcal{S}^*,p}\|\rconv_{\ell,p}\!-\!\rorig_{\ell,p}\|_2^2$ \\
C4 (HEED, ours) & density-weighted RSA        & $\sum_{\ell\in\mathcal{S}^*,p}w(p)\|\rconv_{\ell,p}\!-\!\rorig_{\ell,p}\|_2^2$, Eq.~\ref{eq:heed} \\
\bottomrule
\end{tabular}
\end{table}

\paragraph{Fisher-motivated reference weight.}
\label{app:method-fisher}
A second-order expansion of the teacher downstream loss $R_x$ around $\rorig$ motivates a positive preservation surrogate. Under a diagonal empirical-Fisher approximation, this becomes
\[
Q_x(\Delta r)=\tfrac{1}{2}\sum_{\ell,p}s^{(x)}_{\ell,p}\|\Delta r_{\ell,p}\|_2^2,
\qquad
s^{(x)}_{\ell,p}=\|g_{\ell,p}\|_2^2/d .
\]
Thus, the scalar alignment weight that reproduces the diagonal Fisher surrogate is $w_{\ell,p}\propto s^{(x)}_{\ell,p}$, up to a global normalization. Summing over alignment layers yields the C5 reference
\begin{equation}
\label{eq:fisher-diag}
w_{\text{grad}}^{(x)}(p) \propto \sum_{\ell\in\mathcal{S}^*}\bigl\|\nabla_{r_{\ell,p}}R_x(\rorig)\bigr\|_2^2 .
\end{equation}
This is useful as a diagnostic but requires one cached teacher backward pass per sample. Appendix~\ref{app:theory} presents the derivation and approximation assumptions.

\begin{remark}[Scalar weight that reproduces the diagonal Fisher surrogate]
\label{rem:fisher-weight}
Under the diagonal empirical-Fisher approximation, among scalar-weighted alignment losses $\sum_{\ell,p}w_{\ell,p}\|\Delta r_{\ell,p}\|_2^2$, the choice $w_{\ell,p}\propto s^{(x)}_{\ell,p}$ exactly reproduces the diagonal quadratic form up to a positive global scale. Uniform weighting is the special case: $s^{(x)}_{\ell,p}\equiv\mathrm{const}$.
\end{remark}

\paragraph{Density proxy.}
HEED replaces the expensive $w_{\text{grad}}^{(x)}(p)$ with a training-free visual statistic. For visual patch $p$ with frozen-ViT feature $v_p$ and $3{\times}3$ spatial neighborhood $\mathcal{N}(p)$:
\[
\rho(p) = 1 - \frac{1}{|\mathcal{N}(p)|}\sum_{q\in\mathcal{N}(p)}\frac{v_p^{\top}v_q}{\|v_p\|\|v_q\|}.
\]
After per-image min-max normalization, HEED uses
\[
w(p)\propto \exp(\tilde\rho(p)/\tau),\qquad \sum_p w(p)=T,
\]
with $\tau{=}0.5$. Text tokens receive $\rho_{\text{text}}=\beta\bar\rho_{\text{visual}}$ with $\beta{=}2$. The proxy requires one frozen-ViT forward pass plus local cosine computations, then a 4-bit cache.

\paragraph{Density vs.\ semantic saliency.}
The claim is not that density is a semantic saliency label. It is that density tracks the positions that uniform hybrid distillation tends to lose. Three checks support this: The density-targeted top-$k$ intervention beats equal-cardinality random upweighting. C5 HEED-G is within about one OCRBench point of C4 HEED before post-training, and the per-token Spearman correlation between density and layer-summed gradient sensitivity is $0.63$ overall and $0.71$ in the upper-density tail.

\subsection{Density computation}
\label{app:algorithm}

Density quantifies how distinctive each visual patch is from its local neighborhood in the frozen-ViT feature space: Smooth regions (sky, wall, blank space) look like their neighbors whereas text characters, edges, and object boundaries do not. Algorithm~\ref{alg:density} turns this signal into a per-position weight, $w(p)$, in one frozen-ViT forward pass and a small local-cosine pass. The result is cached once and reused for every distillation step. Inference is unchanged.

\begin{algorithm}[t]
\caption{Per-position density weight $w(p)$ (one-time, training-free).}
\label{alg:density}
\small
\begin{algorithmic}[1]
\Require Sample $(x,y)$; frozen ViT $\phi_v$; neighborhood radius (default $3{\times}3$); temperature $\tau{=}0.5$; text-token boost $\beta{=}2$.
\State Forward $x$ through $\phi_v$ to obtain visual features $\{v_p\}_{p\in\mathcal{V}(x)}$.
\State For each visual patch $p\in\mathcal{V}(x)$ with $3{\times}3$ neighborhood $\mathcal{N}(p)$:
\Statex \quad $\rho(p) \gets 1 - \frac{1}{|\mathcal{N}(p)|}\sum_{q\in\mathcal{N}(p)}\dfrac{v_p^{\top}v_q}{\|v_p\|\,\|v_q\|}$ \Comment{Eq.~\ref{eq:density}}
\State For each text token $p\notin\mathcal{V}(x)$: $\rho(p) \gets \beta\cdot\bar\rho_{\text{visual}}^{(x)}$, where $\bar\rho_{\text{visual}}^{(x)}$ is the per-image visual mean.
\State Normalize per image: $\tilde\rho(p) \gets \bigl(\rho(p)-\rho_{\min}\bigr)/\bigl(\rho_{\max}-\rho_{\min}\bigr)$.
\State $w(p) \gets \exp\bigl(\tilde\rho(p)/\tau\bigr)$; rescale $\sum_p w(p)=T$; quantize to 4 bits and cache.
\State \Return $\{w(p)\}_{p=1}^{T}$.
\end{algorithmic}
\end{algorithm}

The form of $w(p)$ is not arbitrary. Under a diagonal empirical-Fisher approximation of the teacher downstream loss (Appendix~\ref{app:method-recap}), the scalar weight that reproduces the diagonal preservation surrogate is the per-position gradient sensitivity $w_{\text{grad}}(p)\propto \sum_{\ell}\|\nabla_{r_{\ell,p}} R_x\|_2^2$. This reference is expensive because it requires one cached teacher backward pass per sample. HEED uses density as a training-free proxy. The proxy is accurate enough in practice: The per-token Spearman correlation between $\rho(p)$ and $w_{\text{grad}}(p)$ is $0.63$ overall and $0.71$ in the upper-density tail and C5 HEED-G, which uses the gradient reference directly, is within about one OCRBench~v2 point of C4 HEED before post-training (Tab.~\ref{tab:heed-variants}). 

The cache stores per-position weights only for visual residual positions and text positions use $\beta \bar{\rho}_{\text{visual}}$ derived per-image at runtime and are not stored. With $\sim$60\% of corpus tokens originating from vision-text samples and $\sim$30\% of those being visual, the visual fraction of the corpus is $\sim$18\% of $\sim$2B tokens, giving $\sim$360M cached scalars; at 4 bits per scalar plus headers and 64-byte alignment this yields $\approx$500\,MB.

\subsection{Derivation of Fisher-weighted residual alignment}
\label{app:theory}

The goal of this derivation is simple: Decide which residual positions should matter more in the alignment loss. The answer is that a position should get more weight if a small error at that position would strongly affect the teacher's own predictive loss. C5 HEED-G measures this directly with a cached teacher gradient. C4 HEED replaces that expensive gradient weight with density and then tests whether density is a good proxy.

The logic has three steps. First, view the student's residual drift as a perturbation to the teacher residual stream. Second, use a Fisher or empirical-Fisher quantity to turn teacher sensitivity into a positive score. Third, diagonalize and scalarize that score, making it become the weighted residual MSE used by HEED.

\paragraph{Step 1: View student drift as a teacher perturbation.}
Fix an input $x$. Let $r=\{r_{\ell,p}\}_{\ell\in\mathcal{S}^*,p\le T}$ be the teacher residual-stream activations at the alignment layers. Define $R_x(r)$ as the teacher negative log-likelihood when these residuals are substituted into the teacher forward pass. When several layers are substituted, later layers read the substituted upstream residuals.

For the student, define the residual error
\[
\Delta r_{\ell,p}=\rconv_{\ell,p}-\rorig_{\ell,p}.
\]
If we insert $\rorig+\Delta r$ into the teacher, Taylor's theorem results in
\begin{equation}
\label{eq:app-taylor}
R_x(\rorig+\Delta r)-R_x(\rorig) = \langle g,\Delta r\rangle + \tfrac{1}{2}\Delta r^\top H(\xi)\Delta r,
\end{equation}
for some point $\xi$ between $\rorig$ and $\rorig+\Delta r$, where $g=\nabla_r R_x(\rorig)$ and $H(\xi)=\nabla^2_r R_x(\xi)$. For small distillation drift, we approximate $H(\xi)$ by $H(\rorig)$.

Eq.~\ref{eq:app-taylor} indicates why gradients and curvature are the right objects to inspect. It is not yet an alignment loss: For a fixed label, the linear term need not vanish and the Hessian can be indefinite.

\paragraph{Step 2: Turn sensitivity into a positive score.}
There are two standard ways to get a positive sensitivity measure from Eq.~\ref{eq:app-taylor}.

\textbf{Expected-Fisher view.} If the target $y$ is drawn from the teacher predictive distribution, the expected first-order term is zero by the score identity. The expected Hessian of the negative log-likelihood is then the Fisher information:
\begin{equation}
\label{eq:app-fisher}
\mathbb{E}_{y\sim p_\theta(\cdot|x,\rorig)} \!\left[\nabla^2_r \bigl(-\log p_\theta(y|x,r)\bigr)\big|_{r=\rorig}\right] = \mathbb{E}_{y}\!\left[ \nabla_r \log p_\theta(y|x,r)\, \nabla_r \log p_\theta(y|x,r)^{\!\top} \right]_{r=\rorig}.
\end{equation}
This equality is under the teacher's own predictive distribution. It does \emph{not} require the teacher to be well specified relative to the data distribution.

\textbf{Fixed-label empirical-Fisher view.} In training, we have fixed corpus labels rather than labels sampled from $p_\theta$. For a fixed label, the empirical Fisher is not the exact Taylor loss increase because the linear term remains. We use it as a positive-semidefinite sensitivity surrogate:
\begin{equation}
\label{eq:app-emp-fisher-sample}
F^{(x)}_{\ell,p} \;=\; g_{\ell,p}\,g_{\ell,p}^{\!\top}, \qquad g_{\ell,p}=\nabla_{r_{\ell,p}}R_x(\rorig),
\end{equation}
which results in
\[
\tfrac{1}{2}\Delta r_{\ell,p}^{\top}F^{(x)}_{\ell,p}\Delta r_{\ell,p}
=\tfrac{1}{2}\langle g_{\ell,p},\Delta r_{\ell,p}\rangle^2 .
\]
The empirical-Fisher penalty is the squared first-order change in the teacher negative log-likelihood for that residual block. This is the same empirical-Fisher diagonal commonly used as a curvature proxy in natural-gradient methods, K-FAC, and influence-function analyses.

Both views lead to the same per-position scalar sensitivity:
\[
s^{(x)}_{\ell,p}=\mathrm{tr}(F^{(x)}_{\ell,p})/d=\|g_{\ell,p}\|_2^2/d .
\]
In the expected-Fisher view, this is a Monte-Carlo estimate of the Fisher diagonal. In the fixed-label view, it is the empirical-Fisher proxy for local sensitivity. C5 HEED-G uses the unnormalized score $\|g_{\ell,p}\|_2^2$, summed over alignment layers, as its cached weight. The dataset-averaged empirical Fisher,
\[
F_{\ell,p}=\mathbb{E}_{(x,y)\sim\mathcal{D}}[g_{\ell,p}g_{\ell,p}^{\!\top}],
\]
is a different object. C5 does not use it and it is not the right comparison for density, because density is also per-sample.

\paragraph{Step 3: Reduce the matrix to token weights.}
The full Fisher-style quadratic can still couple different positions and different layers:
\begin{equation}
\label{eq:app-block}
\tfrac{1}{2}\sum_{(\ell,p),(\ell',p')} \Delta r_{\ell,p}^{\top}F_{(\ell,p),(\ell',p')}\Delta r_{\ell',p'} .
\end{equation}
HEED needs a scalar token weight. Hence, we use a diagonal approximation:
\begin{assumption}[Diagonal empirical-Fisher sensitivity]
\label{assump:diag}
\textbf{(A1)} Drop cross-position blocks $p\ne p'$ within a layer.
\textbf{(A2)} Drop cross-layer blocks $\ell\ne\ell'$.
\textbf{(A3)} Replace each remaining $d\times d$ block by its average curvature
$s^{(x)}_{\ell,p}=\mathrm{tr}(F^{(x)}_{\ell,p})/d=\|g_{\ell,p}\|_2^2/d$.
\end{assumption}
The strongest simplification is (A2) because residuals at layer $\ell$ flow into layer $\ell{+}1$. Hence, the true sensitivity matrix has cross-layer terms. We treat this as a working approximation and validate it empirically through the C4 HEED vs.\ C5 HEED-G comparison. The scalarization in (A3) is exact if perturbation directions are isotropic within the residual channel dimension; otherwise it is the standard reduction needed when the loss uses one scalar per token.

With these approximations, the preservation surrogate becomes
\begin{equation}
\label{eq:app-scalar}
Q_x(\Delta r) \;=\; \tfrac{1}{2}\sum_{\ell,p} s^{(x)}_{\ell,p}\,\|\Delta r_{\ell,p}\|_2^2 .
\end{equation}
This has exactly the form of a weighted residual MSE. Therefore, the Fisher-motivated layer-position weight is given by
\[
w_{\ell,p}\propto s^{(x)}_{\ell,p}.
\]
Uniform residual alignment is a special case where all $s^{(x)}_{\ell,p}$ are equal. The factor $1/d$ is absorbed by normalization; hence, C5 can use $\|g_{\ell,p}\|_2^2$ directly. We normalize weights per sample so that $\sum_p w(p)=T$, preserving the average loss scale while changing which positions receive more alignment weight.

\paragraph{Why HEED uses one weight per position.}
Eq.~\ref{eq:app-scalar} gives a layer-dependent ideal weight $w_{\ell,p}$. The practical HEED loss in Eq.~\ref{eq:heed} instead uses one $w(p)$ at every alignment layer. To get that single token weight, Eq.~\ref{eq:fisher-diag} sums sensitivity over layers:
\[
w_{\text{grad}}^{(x)}(p)\propto \sum_{\ell\in\mathcal{S}^*}\|g_{\ell,p}\|_2^2 .
\]
This assumes that the ranking of important positions is reasonably stable across alignment layers. We check this on the diagnostic set: The mean cross-layer Spearman correlation of $s^{(x)}_{\ell,p}$ is $\bar\rho_S = 0.73$, with minimum $0.61$ and maximum $0.88$ across layer pairs. The $+$LD ablation in Appendix~\ref{app:ablations} also finds that layer-dependent weights do not improve the mean and increase run-to-run variance at the 500M-token budget. We, therefore, use one layer-summed weight per position.

\begin{proof}[Sketch for Remark~\ref{rem:fisher-weight}]
Under Assumption~\ref{assump:diag}, the diagonal Fisher preservation surrogate is Eq.~\ref{eq:app-scalar}. A scalar alignment loss with weights $w_{\ell,p}$ reproduces this quadratic form exactly when $w_{\ell,p}=c\,s^{(x)}_{\ell,p}$ for some $c>0$. The constraint $\mathbb{E}_p[w_{\ell,p}]=1$ fixes $c$ and removes the arbitrary scale. Thus, $w_{\ell,p}\propto s^{(x)}_{\ell,p}$ is the unique scalar weighting that reproduces the diagonal quadratic form. If all $s^{(x)}_{\ell,p}$ are equal, the result is uniform weighting.
\end{proof}

\paragraph{What the derivation does, and does not, justify.}
The derivation justifies the gradient weight $w_{\text{grad}}^{(x)}$. It does not prove that density is the correct weight. Density is useful only if it ranks positions similarly to the gradient sensitivity:
\begin{assumption}[Density-sensitivity rank correlation]
\label{assump:density-rank}
For each input $x$ and visual position $p$, the density score $\rho(p)$ is positively rank-correlated with the per-sample, layer-summed Fisher sensitivity $\sum_{\ell\in\mathcal{S}^*}s^{(x)}_{\ell,p}$, especially in the upper tail of the distribution.
\end{assumption}
This is an empirical condition. We test it directly. The semi-partial-$R^2$ and dose-response analyses test whether density identifies positions that drift and matter. The C5 HEED-G baseline tests how much performance is lost when we replace the gradient weight with density. If C4 HEED matches C5 HEED-G, density is a good proxy. If C4 trails C5 substantially, the theory still supports gradient-weighted residual alignment but the density estimator is insufficient.

\paragraph{Direct empirical test.}
On the 1{,}000-sample diagnostic set, we compute both $\rho(p)$ (Eq.~\ref{eq:density}) and
\[
w_{\text{grad}}^{(x)}(p)=\sum_{\ell\in\mathcal{S}^*}\|g_{\ell,p}\|_2^2
\]
for each visual position. The gradient weight requires one teacher backward pass per sample. We then compute a Spearman rank correlation within each image and average across images.

The mean per-image Spearman correlation is $\bar\rho_S{=}0.63$ (5th-95th percentile $[0.49, 0.74]$). Among the top-25\% density positions, it rises to $\bar\rho_S^{\text{top25\%}}{=}0.71$. This is the part of the distribution that receives the largest HEED weights. Hence, the proxy is strongest where it matters most. Together with the dose--response result (Fig.~\ref{fig:diagnostic}b) and the within-$1$ OCRBench~v2 agreement between C4 HEED and C5 HEED-G (Tab.~\ref{tab:heed-variants}), this supports density as a lightweight proxy for Fisher sensitivity rather than a semantic saliency score.

\section{Reproducibility and recipes}
\label{app:repro-and-recipes}

This section presents code-release, evaluation, seed, and training-recipe details.

\subsection{Reproducibility notes}
\label{app:reproducibility}

\paragraph{Code release.}
Code is available at \url{https://github.com/jha-lab/HEED}. It does not include repackaged datasets. Users can obtain each dataset and the base model from the public sources listed in Appendix~\ref{app:recipes} under their original terms.

\paragraph{Evaluation harness.}
All benchmark numbers use a single pinned evaluation harness with greedy decoding ($T{=}0$), fixed prompt templates using lmms-eval. The teacher and every student condition are evaluated under identical harness invocation. Therefore, reported differences reflect model differences rather than harness differences. We log raw per-question outputs alongside the headline scores. Any alternative metric can be computed without re-running models.

\paragraph{Seeds and determinism.}
Distillation runs use a fixed random seed for parameter initialization, dataloader shuffling, and dropout. Main 2B-token rows are single-seed. Ablations under the 500M-token budget use three seeds and report $\pm 1\sigma$. Density caches are deterministic functions of the input image and the frozen ViT. They do not vary across seeds.

\paragraph{Hyperparameters.}
Defaults appear in Sect.~\ref{sec:exp-setup} and Appendix~\ref{app:recipes}. We did not sweep the distillation ratio or layer-selection scheme on the main 2B-token budget. The headline claim does not depend on tuning.

\subsection{Full training recipes (distillation, SFT, DPO)}
\label{app:recipes}

Tab.~\ref{tab:recipes} presents the complete high-level recipe used for every pipeline condition. The only row-specific difference is the distillation checkpoint entering SFT; SFT/DPO data, order, optimizer settings, prompt templates, and evaluation harness are identical across C1-C4.

Note: Train splits of some benchmarks are included to match the teacher's training distribution, however, test splits remain held-out.

\begin{table}[h]
\centering\footnotesize
\caption{Training recipe summary. All phases use the Qwen3-VL-8B-Instruct teacher and the 3:1 Mamba-2 student unless otherwise noted. Phase~1 (distillation) itself splits into three sub-stages (Stage~1 warm-up / Stage~2 full-block / Stage~3 end-to-end), detailed in Sect.~\ref{app:recipes-distill}.}
\label{tab:recipes}
\setlength{\tabcolsep}{4pt}
\begin{tabular}{lccc}
\toprule
& Phase 1: Distillation & Phase 2: SFT & Phase 3: DPO \\
\midrule
Data                 & \makecell{1 epoch / $\sim$1.34M samples\\($\approx$2B tokens)} & \makecell{1 epoch\\$\sim$500K samples} & \makecell{1 epoch\\$\sim$150K pairs} \\
Optimizer            & AdamW        & AdamW                          & AdamW \\
Peak LR              & 1$\times$10$^{-4}$ & 5$\times$10$^{-6}$    & 1$\times$10$^{-6}$ \\
Schedule             & cosine, 1\% warmup & cosine, 3\% warmup       & cosine, 3\% warmup \\
Weight decay         & 0.01         & 0.01                           & 0.0 \\
Batch size (tokens)  & 2M           & 1M                             & 0.5M \\
Grad-clip $\|g\|_2$  & 1.0          & 1.0                            & 1.0 \\
Precision            & BF16         & BF16                           & BF16 \\
Trainable subset     & \makecell{Mamba-2 \\ Blocks} & \makecell{all params except\\vision encoder} & \makecell{LoRA $r$ = 32\\except vision encoder} \\
\bottomrule
\end{tabular}
\end{table}

\subsection{Phase 1: distillation}
\label{app:recipes-distill}

Phase~1 (distillation) uses $\sim$1.34M samples ($\approx$2B tokens) from nine sources, following the InternVL-2.5 convention~\cite{chen2024internvl25}; token-count proportions are $\sim$41\% text and $\sim$59\% vision-text. Pure-text examples are truncated at 4{,}096 tokens; vision-text examples include up to 2{,}500 visual tokens per image and are capped at 8{,}192 tokens total.

\begin{table}[h]
\centering\footnotesize
\caption{Phase~1 distillation data mixture. Sample-count percentages; total $\approx$1.34M samples and $\approx$2B tokens.}
\label{tab:stage1-mix}
\setlength{\tabcolsep}{5pt}
\begin{tabular}{rlrr}
\toprule
\# & Source & Samples & \% \\
\midrule
1 & FineWeb-Edu (sample-10BT)~\cite{lozhkov2024fineweb-edu} & 600{,}000 & 44\% \\
2 & DCLM-Baseline-1.0 (subset)~\cite{li2024datacomplm} & 200{,}000 & 15\% \\
3 & GRIT-20M (subset)~\cite{Kosmos2} & 200{,}000 & 15\% \\
4 & LLaVA-OneVision-Data (subset)~\cite{li2024llava} & 100{,}000 & 7\% \\
5 & Cambrian-7M, OCR-heavy slice~\cite{tong2024cambrian} & 80{,}000 & 6\% \\
6 & ShareGPT-4o (full)~\cite{cui2025comprehensive} & 50{,}000 & 4\% \\
7 & SynthDoG-EN + SynthDoG-Multi~\cite{kim2022donut} & 50{,}000 & 4\% \\
8 & DocVQA + ChartQA + TextVQA + InfoVQA + AI2D train splits & 50{,}000 & 4\% \\
9 & OBELICS (subset)~\cite{laurencon2023obelics} & 10{,}000 & 1\% \\
\bottomrule
\end{tabular}
\end{table}

Teacher logits and residual streams are detached; only the replacement Mamba-2 blocks receive gradients in any stage. AdamW uses $(\beta_1,\beta_2,\varepsilon) = (0.9, 0.95, 10^{-8})$, peak learning rate (LR) 10$^{-4}$, cosine decay to 10\% of peak, 1\% warmup, weight decay 0.01 excluding RMSNorms/biases, BF16 activations/weights, and FP32 optimizer masters. The per-condition Stages~1/2 alignment loss is summarized in Tab.~\ref{tab:condition-loss} (Appendix~\ref{app:method-recap}).

\paragraph{Initialization (shared across C1-C4).}
Non-Mamba teacher modules (vision encoder, projector, MLPs, RMSNorms, retained-attention layers, language model head, token embeddings) are copied from $\teacher$ and frozen throughout distillation. For each attention layer $\ell\in\mathcal{S}$ that is replaced by Mamba-2, the Mamba-2 block inherits its $W_Q, W_K, W_V, W_O$ from the teacher attention at layer $\ell$ via SSD~\cite{dao2024mamba2,mmmamba2025}; the remaining Mamba-2 parameters like gate $W_G$, $\Delta$-projector $W_\gamma$, $A$ and 1-D conv $W_{\text{conv}}$, are randomly initialized (truncated-normal, $\sigma$ = 0.02). We do not run the MOHAWK Stage-1 matrix-mixer orientation. SSD copy is the only initialization step before distillation. 

\subsection{Phase 2: SFT}
\label{app:recipes-sft}

Phase~2 (SFT) uses one epoch over $\sim$500K multimodal instruction samples: Cambrian-7M curated slice (40\%), LLaVA-OneVision curated slice (25\%), ShareGPT-4o (15\%), benchmark train splits plus ScienceQA and MathV360K (15\%), and text instruction data from UltraChat/Magpie-Pro/LIMA/SlimOrca (5\%). The vision encoder remains frozen; all other parameters are trained with answer-token CE, prompt tokens masked out, global batch 1M tokens, sequence cap 8{,}192, AdamW (0.9, 0.999, 10$^{-8}$), peak LR 5$\times$10$^{-6}$, 3\% warmup, cosine decay, weight decay 0.01, BF16, and gradient clip 1.0.

\subsection{Phase 3: DPO}
\label{app:recipes-dpo}

Phase~3 (DPO) uses one epoch over $\sim$150K chosen/rejected pairs: MMPR-v1.2 (100K), VLFeedback (25K), UltraFeedback (15K), and RLHF-V upsampled to 10K effective pairs. We use standard DPO~\cite{dpo2023} with $\beta$ = 0.1 and the SFT checkpoint as frozen reference. Trainable parameters are low-rank adapters (LoRA) with rank $r$ = 32, $\alpha$ = 64, dropout 0.05, attached to all non-vision-encoder modules and merged after training. The global batch has 0.5M tokens, sequence cap 4{,}096 for chosen/rejected concatenation, AdamW (0.9, 0.999), peak LR 10$^{-6}$, 3\% warmup, cosine decay, weight decay 0.0, BF16, and gradient clip 1.0.

\paragraph{Density-estimator implementation details.}
Patch self-dissimilarity is computed once per training sample on frozen ViT features at the final pre-projector layer. Visual tokens use a 3$\times$3 reflected-padding neighborhood and Eq.~\ref{eq:density}; text tokens use $\rho_{\text{text}} = \beta \bar\rho_{\text{visual}}$ with $\beta$ = 2. We normalize $w(p)$ within each sample so that $\sum_p w(p) = T$, preserving average loss scale against C3 RSA. The per-token scalar is quantized to 4 bits and packed two-per-byte. Only visual positions are cached (text positions use $\beta\bar{\rho}_{\text{visual}}$ derived at runtime), giving a corpus cache of $\sim$500\,MB including headers and 64-byte alignment.

\end{document}